\documentclass[11pt,a4paper]{article}

\newcommand{\ReportTitle}{KineWorld: Action-Induced Transport Fields for Embodied World Modeling}
\newcommand{\ReportShortTitle}{KineWorld}
\newcommand{\ReportAuthors}{%
\mbox{Ziying Song\textsuperscript{1,*}}, \mbox{Yuchen Liu\textsuperscript{2,*}}, \mbox{Zhuoran Xu\textsuperscript{3,*}}, \mbox{Ziyang Liu\textsuperscript{4}}, \mbox{Jian Jin\textsuperscript{5}},
\mbox{Jiangtao Su\textsuperscript{1}}, \mbox{Haibao Yu\textsuperscript{6}}, \mbox{Lei Yang\textsuperscript{1,\ensuremath{\dagger}}}, \mbox{Yuanpei Chen\textsuperscript{7}}%
}
\newcommand{\ReportAffiliations}{%
\mbox{\textsuperscript{1}Nanyang Technological University}\quad
\mbox{\textsuperscript{2}North University of China}\quad
\mbox{\textsuperscript{3}Psibot}\quad
\mbox{\textsuperscript{4}The Hong Kong Polytechnic University}\quad
\mbox{\textsuperscript{5}China Academy of Information and Communications Technology}\quad
\mbox{\textsuperscript{6}University of Hong Kong}\quad
\mbox{\textsuperscript{7}Peking University}%
}
\newcommand{\ReportAuthorNotes}{\textsuperscript{*}Co-first authors\quad \textsuperscript{\ensuremath{\dagger}}Corresponding author}
\newcommand{\ReportDate}{October 2026}
\newcommand{\ReportPDFAuthors}{Ziying Song, Yuchen Liu, Zhuoran Xu, Ziyang Liu, Jian Jin, Jiangtao Su, Haibao Yu, Lei Yang, Yuanpei Chen}

\usepackage[left=25mm,right=25mm,top=18mm,bottom=20mm,
  headheight=14pt,headsep=10pt]{geometry}
\usepackage[T1]{fontenc}
\usepackage{mathptmx}
\usepackage[scaled=0.94]{helvet}
\usepackage{inconsolata}
\usepackage{microtype}
\usepackage{amsmath,amssymb,bm}

\usepackage{amsmath,amsfonts,bm}

\def\eqref#1{equation~\ref{#1}}

\def\1{\bm{1}}

\DeclareMathAlphabet{\mathsfit}{\encodingdefault}{\sfdefault}{m}{sl}
\SetMathAlphabet{\mathsfit}{bold}{\encodingdefault}{\sfdefault}{bx}{n}

\usepackage{xspace}
\usepackage{docmute}
\usepackage{graphicx}
\usepackage{float}
\usepackage{array,booktabs,tabularx,multirow,makecell}
\usepackage{algorithm,algorithmic}
\usepackage{enumitem}
\usepackage{placeins}
\usepackage{ragged2e}
\usepackage[table]{xcolor}
\usepackage[most]{tcolorbox}
\usepackage[authoryear,round]{natbib}
\usepackage{titlesec}
\usepackage{fancyhdr}
\usepackage{caption}
\usepackage{url}
\usepackage{hyperref}
\usepackage[capitalize]{cleveref}

\definecolor{reportbackground}{HTML}{F3F0F8}
\definecolor{reportaccent}{HTML}{62438D}
\definecolor{reportmuted}{HTML}{776A87}
\definecolor{reportink}{HTML}{000000}
\hypersetup{
  colorlinks=true,urlcolor=reportaccent,citecolor=reportaccent,
  linkcolor=reportaccent,
  pdftitle={\ReportTitle},pdfauthor={\ReportPDFAuthors},
  pdfsubject={Technical Report}
}

\titleformat{\section}{\large\bfseries}{\thesection}{0.8em}{}
\titleformat{\subsection}{\normalsize\bfseries}{\thesubsection}{0.8em}{}
\titleformat{\subsubsection}{\normalsize\bfseries}{\thesubsubsection}{0.8em}{}
\titleformat{\paragraph}[runin]{\normalsize\bfseries}{}{0pt}{}
\titlespacing*{\section}{0pt}{16pt}{7pt}
\titlespacing*{\subsection}{0pt}{12pt}{5pt}
\titlespacing*{\subsubsection}{0pt}{10pt}{4pt}
\titlespacing*{\paragraph}{0pt}{7pt}{0.6em}
\renewcommand{\bibfont}{\small}

\crefname{section}{Sec.}{Secs.}
\Crefname{section}{Sec.}{Secs.}
\crefname{appendix}{Appendix}{Appendices}
\Crefname{appendix}{Appendix}{Appendices}
\crefname{table}{Tab.}{Tabs.}
\Crefname{table}{Table}{Tables}
\crefname{equation}{Eq.}{Eqs.}
\Crefname{equation}{Eq.}{Eqs.}
\crefname{figure}{Fig.}{Figs.}
\Crefname{figure}{Fig.}{Figs.}

\renewcommand{\headrulewidth}{0pt}
\renewcommand{\footrulewidth}{0pt}
\fancypagestyle{reportfirst}{
  \fancyhf{}
  \fancyfoot[C]{\fontsize{9}{11}\selectfont\color{reportmuted}\thepage}
  \renewcommand{\headrulewidth}{0pt}
  \renewcommand{\footrulewidth}{0pt}
}

\newtcolorbox{reportfront}{
  enhanced,frame hidden,colback=reportbackground,coltext=reportink,
  arc=9pt,boxrule=0pt,boxsep=0pt,
  left=14pt,right=14pt,top=14pt,bottom=13pt,
  before skip=0pt,after skip=0pt
}

\newtcolorbox[auto counter,number within=section]{reportbox}[2]{
  enhanced,breakable,
  colback=reportbackground,colframe=reportaccent,
  colbacktitle=reportaccent,coltitle=white,coltext=reportink,
  fonttitle=\small\bfseries,
  fontupper=\fontsize{8.5}{10.5}\selectfont\rmfamily,
  before upper={\setlength{\parindent}{0pt}\setlength{\parskip}{2pt}},
  sharp corners,boxrule=0.4pt,boxsep=0pt,
  left=8pt,right=8pt,top=5pt,bottom=6pt,
  toptitle=3pt,bottomtitle=3pt,before skip=7pt,after skip=7pt,
  title={Box~\thetcbcounter.\ #1},label={#2}
}

\newif\ifVThreeAppendixStandalone
\VThreeAppendixStandalonefalse
\newcommand{\TAWD}{\textsc{TAWD}\xspace}
\DeclareRobustCommand{\reviewold}[1]{\textcolor{blue}{#1}}
\DeclareRobustCommand{\reviewedit}[1]{\textcolor{red}{#1}}
\DeclareRobustCommand{\KineWorldColor}{KineWorld}
\DeclareRobustCommand{\KineWorldTitleColor}{KineWorld}
\begin{document}
\thispagestyle{reportfirst}
\begin{reportfront}
{\sffamily\fontsize{8}{10}\selectfont\color{reportmuted}
\MakeUppercase{Technical Report}\hfill\ReportDate\par}
\vspace{7pt}
{\sffamily\bfseries\fontsize{21.5}{24}\selectfont\raggedright\color{reportaccent}
\ReportTitle\par}
\vspace{6pt}
{\sffamily\bfseries\fontsize{9.5}{11.5}\selectfont\raggedright\ReportAuthors\par}
\vspace{3pt}
{\fontsize{8.5}{10.5}\selectfont\raggedright\ReportAffiliations\par}
\vspace{3pt}
{\fontsize{8.5}{10.5}\selectfont\ReportAuthorNotes\par}
\vspace{9pt}
{\fontsize{10.5}{12.7}\selectfont\justifying
\setlength{\parindent}{0pt}
Embodied world models predict the visual consequences of candidate actions before execution. However, existing action-conditioned world models often adopt uniformly weighted visual generation objectives that can be misaligned with embodied prediction needs. Even with explicit motion conditioning, these objectives can underemphasize spatially sparse changes that are critical to interaction.
We propose \textbf{\KineWorldTitleColor}, a transport-aware world-modeling framework that extends robot kinematics from motion conditioning to the spatial allocation of generative supervision. \textbf{Kinematic Transport Lifting (KTL)} constructs renderer-derived, camera-aligned transport fields from commanded robot motion. \textbf{Transport-Aware World Diffusion (TAWD)} calibrates their motion support on the video-latent grid and reweights future-RGB flow matching through a normalized mixture of uniform and transport-focused distributions.
We train KineWorld using ALOHA--AgileX bimanual manipulation data from \textbf{RoboTwin~2.0}. KineWorld achieves an \textbf{EWMScore-P of 68.95} in single-view evaluation and a \textbf{TWB-Score of 54.82} in multi-view evaluation. These results support a shift from appearance fitting toward action-consequence modeling for embodied decision-making.
\par}
\vspace{6pt}
{\sffamily\fontsize{9}{11}\selectfont
\textbf{Code:} \url{https://github.com/modaxiansheng/KineWorld}\par
\textbf{Project:} \url{https://modaxiansheng.github.io/KineWorld/}\par}
\end{reportfront}

\suppressfloats[t]

\section{Introduction}
\label{sec:introduction}
World Action Models (WAMs) enable robots to anticipate how the visual world will evolve under candidate actions before executing them in the physical environment. By generating action-conditioned future observations, these models can serve as predictive simulators for policy learning, action evaluation, planning, and synthetic experience generation. Recent advances in large-scale video generation further provide strong appearance and dynamics priors, making it possible to construct visually realistic robot rollouts from limited embodied data \cite{zhu2024irasim,guo2025ctrlworld,liao2025genieenvisioner}.

\begin{figure}[t]
    \centering
    \includegraphics[width=\linewidth]
    {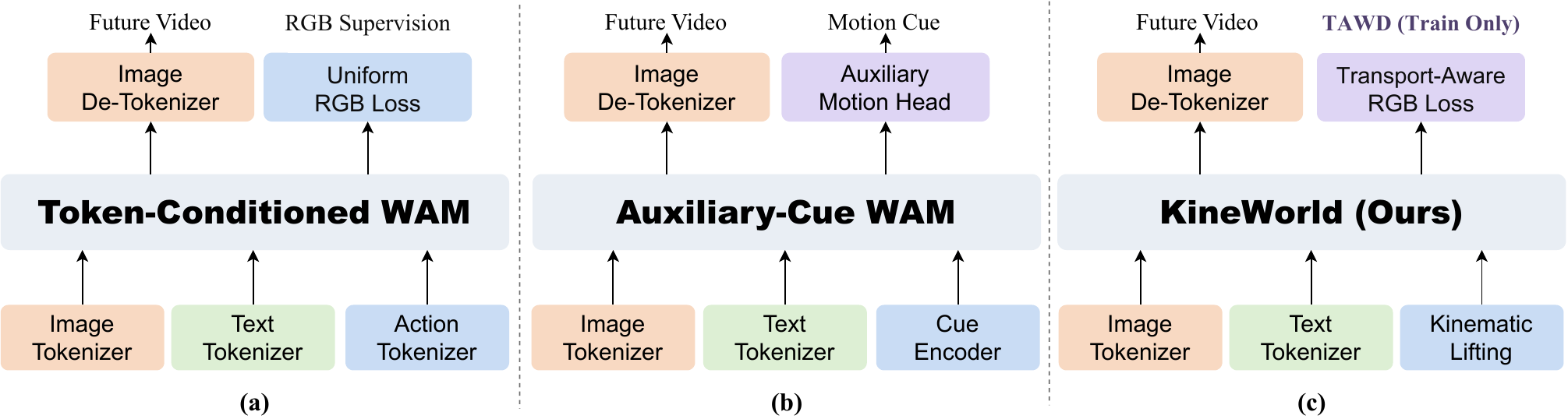}
    \caption{
    \textbf{(a)} Token-conditioned WAMs use numerical trajectories
    or learned action embeddings for video generation
    \citep{zhu2024irasim,guo2025ctrlworld,liao2025genieenvisioner},
    but typically apply \emph{uniform RGB supervision}, potentially
    \textbf{underemphasizing interaction-critical local changes}.
    \textbf{(b)} Auxiliary-cue WAMs incorporate
    \emph{motion-focused auxiliary cues}, such as masks \citep{bridgev2w},
    action images \citep{action_images}, optical flow \citep{chen2026flowwam},
    or projected kinematic fields \citep{eawm}, to improve motion localization.
    These cues alone do not directly change the spatial weighting of the primary
    RGB objective, potentially
    \textbf{leaving motion localization and supervision priorities misaligned}.
    \textbf{(c)} KineWorld introduces \emph{action-guided RGB supervision}
    by lifting camera-aligned kinematic transport onto the video latent
    grid and reweighting flow matching, enabling
    \textbf{action-consequence-oriented supervision} while preserving
    the total token weight.
    }
    \label{fig:motivation}
\end{figure}

Existing world action models (WAMs) improve action responsiveness through numerical action tokens, learned latent codes, and spatially aligned conditions such as embodiment masks, action images, optical flow, and projected kinematic fields \cite{zhu2024irasim,guo2025ctrlworld,wang2026latentwam,liao2025genieenvisioner,chen2026flowwam}. Yet motion conditioning does not itself alter the spatial weighting of the primary RGB-generation objective. When action-responsive regions occupy only a small part of the frame, uniform token averaging can bias optimization toward static appearance. Motion-sensitive auxiliary objectives and flow-branch reweighting address related concerns \cite{zhu2024irasim,chen2026flowwam}. We further use robot kinematics to guide future-RGB supervision, aligning the model's learning priorities with action consequences.
This raises a key question: \textbf{how can robot kinematics guide the learning of action consequences?}

Our key insight is that action conditioning and supervision allocation serve different purposes: the former specifies what motion to generate, whereas the latter determines which prediction errors receive greater weight during training. Robot kinematics provides a geometry-derived prior for this allocation by identifying where commanded robot motion is expected to appear in the camera view, without relying on generated future frames. This suggests using kinematic transport not only as a generation condition, but also as a spatial prior for the primary
video-generation objective.

Building on this insight, we introduce \textbf{\KineWorldTitleColor}, a
transport-aware training framework for action-conditioned world modeling. \textbf{Kinematic Transport Lifting (KTL)} constructs camera-aligned transport, while \textbf{Transport-Aware World Diffusion (TAWD)} calibrates its motion support on the video latent grid and converts the resulting support into a normalized mixture distribution for the RGB flow-matching objective. The resulting objective redistributes supervision while preserving the aggregate token weight, without adding inference parameters or sampling steps. This extends kinematic priors from generation conditions to learning guidance for interaction- and decision-oriented world prediction.
Our main contributions are summarized as follows:
\begin{itemize}

\item \textbf{Action-Consequence-Oriented Learning.}
We distinguish visual extent from interaction importance, motivating
kinematics-guided learning of interaction-critical local changes.

\item \textbf{\KineWorldTitleColor{} Framework.}
We propose \KineWorldTitleColor{}, a transport-aware framework with KTL and TAWD. KTL constructs camera-aligned transport from the commanded rollout, while TAWD reweights RGB flow matching via a normalized mixture on the latent grid, preserving token weight and sampling.

\item \textbf{Benchmark Evaluation.}
Trained on RoboTwin~2.0 ALOHA--AgileX bimanual manipulation data \citep{chen2025robotwin2}, KineWorld achieves an EWMScore-P of 68.95 and a TWB-Score of 54.82 in single-view and multi-view evaluation, respectively. Our framework and evaluations provide the community with a reference for action-consequence-oriented world modeling.

\end{itemize}
\section{Related Work}
\label{sec:related_work}

\paragraph{\textcolor{black}{World Action Models.}}
World Action Models (WAMs) increasingly leverage pretrained video generators to connect visual prediction with robot control \citep{guo2025ctrlworld,liao2025genieenvisioner}. Existing approaches include action-conditioned video prediction \citep{zhu2024irasim,guo2025ctrlworld}, joint video-action generation \citep{unified_video_action}, and action decoding from visual latent representations \citep{liao2025genieenvisioner}. Recent systems further broaden this paradigm. RealWonder connects physical actions to video generation through simulated RGB and flow \citep{liu2026realwonder}, while CLAP integrates heterogeneous action representations for cross-embodiment learning \citep{liu2026clap}. However, advances in generative architectures and action interfaces do not inherently prioritize interaction-critical changes in the primary RGB objective. When these changes occupy few visual tokens, uniform weighting can underemphasize their contribution. KineWorld addresses this complementary problem through kinematics-guided allocation of future-RGB supervision.

\paragraph{\textcolor{black}{Action Representations for WAMs.}}
Existing WAMs encode robot actions as numerical tokens, learned latent codes, or spatially grounded visual cues \citep{zhu2024irasim,latent_action_model,unified_video_action}. Recent methods align actions with the visual domain through embodiment masks \citep{bridgev2w}, multi-view action videos \citep{action_images}, optical flow \citep{chen2026flowwam}, and camera-aligned kinematic fields \citep{eawm}. Beyond conditioning, FlowWAM reweights its flow-branch loss using reference-relative latent deviations \citep{chen2026flowwam}, while EA-WM uses frame-difference supervision to guide cross-stream fusion \citep{eawm}. In contrast, KineWorld uses kinematic transport to directly reweight future-RGB flow matching without introducing auxiliary objectives.

\section{Preliminaries}
\label{sec:preliminaries}

\subsection{Embodied World Modeling}
\label{sec:embodied_world_modeling}

An episode consists of RGB observations $\{I_t\}_{t=0}^{T}$, robot
configurations $\{q_t\}_{t=0}^{T}$, a language instruction $c$, and an
action trajectory $A=\{a_t\}_{t=0}^{T-1}$, where
$I_t\in\mathbb R^{H\times W\times 3}$ and $a_t$ controls the transition
from $t$ to $t+1$. In our bimanual setting, $a_t\in\mathbb R^{14}$ specifies
an absolute 6-DoF end-effector pose target and one gripper command for each
arm. The action is executed through an embodiment-specific controller that
maps the commanded end-effector targets to robot-configuration transitions.
 An embodied world model learns $p_\theta(I_{1:T}\mid I_0,A,c)$, whose
predictions should preserve both visual realism and action fidelity.

\subsection{Conditional Flow Matching}
\label{sec:conditional_flow_matching}

Given a $K$-frame clip $I_{0:K-1}$, a causal, first-frame-preserving video
VAE produces
$z^x=\mathcal E_{\mathrm{VAE}}(I_{0:K-1})
\in\mathbb R^{C_z\times Q\times H_z\times W_z}$.
We partition it as
$z^x=[z^{x,\mathrm{cond}},z^{x,\mathrm{fut}}]$, where
$z^{x,\mathrm{cond}}=z^x_0$ and
$z^{x,\mathrm{fut}}=z^x_{1:Q-1}$.

Let $\tau\sim p(\tau)$, $\epsilon\sim\mathcal N(0,\mathbf I)$, and
$\sigma(\tau)\in[0,1]$ be a differentiable schedule with
$\sigma(0)=0$ and $\sigma(1)=1$. Only future RGB latents are noised:
\begin{equation}
\begin{aligned}
    \tilde z^x_{\tau,\ell}
    &=
    \begin{cases}
        z^x_0,
        & \ell=0,\\
        (1-\sigma(\tau))z^x_\ell+\sigma(\tau)\epsilon_\ell,
        & 1\leq\ell\leq Q-1,
    \end{cases}\\
    v^\star_{\tau,\ell}
    &=
    \dot{\sigma}(\tau)
    (\epsilon_\ell-z^x_\ell),
    \qquad
    1\leq\ell\leq Q-1.
\end{aligned}
\label{eq:flow_path}
\end{equation}
For $\sigma(\tau)=\tau$,
$v^\star_{\tau,\ell}=\epsilon_\ell-z^x_\ell$.
Conditioned on language $c$ and action representation $\xi$, the model
predicts
$\hat v_\theta=v_\theta(\tilde z^x_\tau,\tau\mid c,\xi)$.
The future latent positions are
\begin{equation}
    \mathcal V
    =
    \{(\ell,h,w):
    1\leq\ell\leq Q-1,\,
    1\leq h\leq H_z,\,
    1\leq w\leq W_z\},
    \qquad
    N=|\mathcal V|.
    \label{eq:future_token_set}
\end{equation}
For $i=(\ell,h,w)\in\mathcal V$, the channel-averaged prediction error is
\begin{equation}
    e_i(\theta)
    =
    \frac{1}{C_z}
    \sum_{\chi=1}^{C_z}
    \left(
        \hat v_{\theta,\chi i}
        -
        v^\star_{\tau,\chi i}
    \right)^2,
    \label{eq:token_error}
\end{equation}
and standard conditional flow matching uniformly averages these errors:
\begin{equation}
    \mathcal L_{\mathrm{FM}}
    =
    \mathbb E_{I,c,\xi,\tau,\epsilon}
    \left[
        \frac{1}{N}
        \sum_{i\in\mathcal V}
        e_i(\theta)
    \right].
    \label{eq:uniform_flow_matching}
\end{equation}
The conditioning frame $\ell=0$ remains clean and is excluded from the
loss; generation integrates the learned vector field from $\tau=1$ to
$\tau=0$.

\begin{figure}[t]
    \centering
    \includegraphics[width=\linewidth]
    {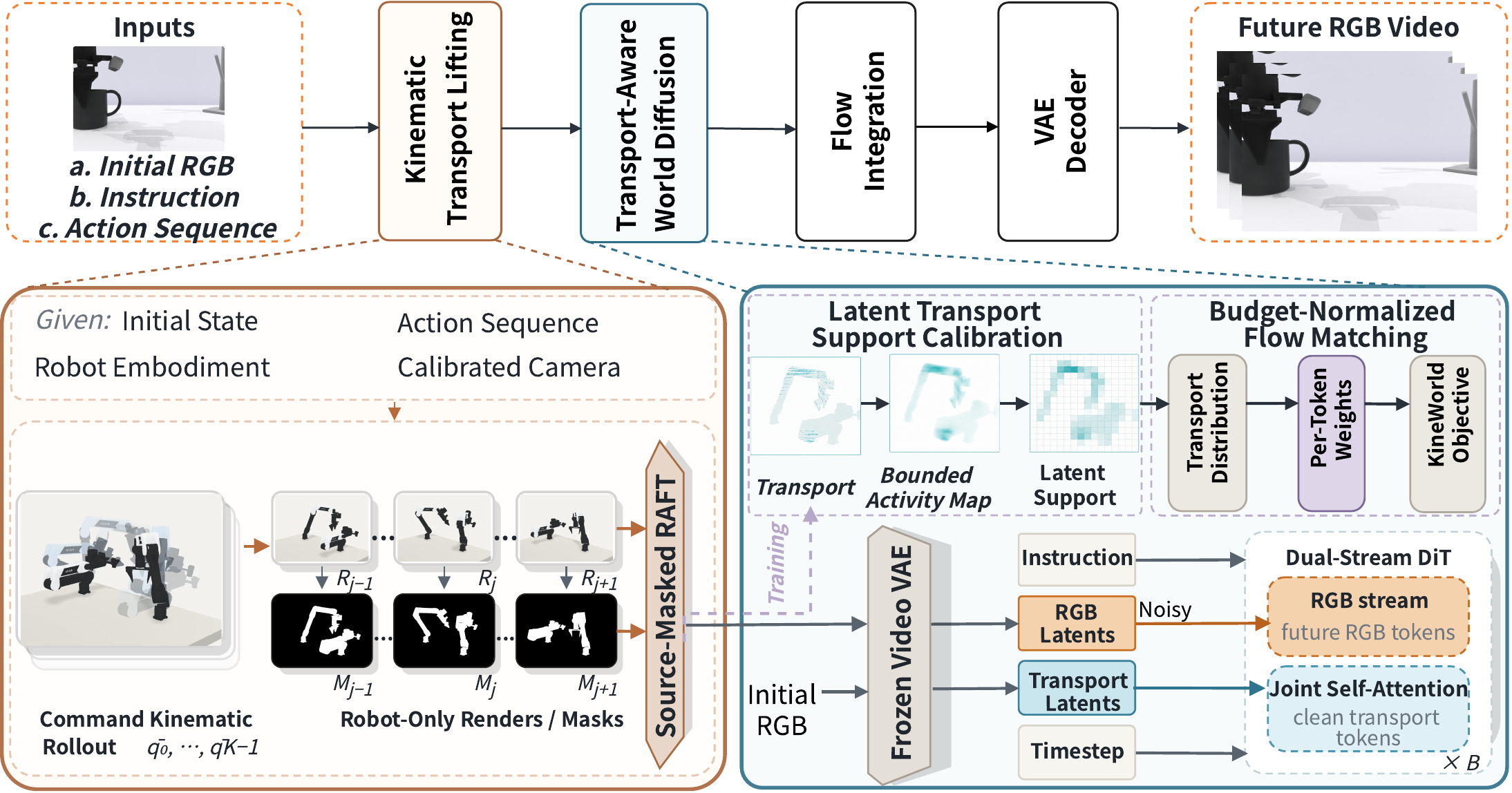}
\caption{\textbf{Overview of KineWorld.}
Kinematic Transport Lifting (KTL) converts the commanded robot rollout into
camera-aligned transport and injects its latent representation as a clean
structural condition. Transport-Aware World Diffusion (TAWD) calibrates the
transport support and derives mean-one token weights for the flow-matching
objective, emphasizing action-responsive regions under a fixed weight budget.
The weighting branch is used only during training, while the inherited action
expert remains unchanged.}
    \label{fig:framework}
\end{figure}
\section{KineWorld}
\label{sec:method}

\subsection{Overview}
\label{sec:overview}

KineWorld is a latent video world model that predicts future RGB observations from the current observation, a language instruction, and a candidate action sequence. A causal video VAE represents visual states in latent space, while a conditional flow model learns their action-conditioned dynamics. When projected robot motion occupies only a small part of the video, uniform token averaging assigns this region a correspondingly small share of the total token weight. KineWorld uses commanded kinematics to guide both generation and learning: camera-aligned transport conditions the predicted motion, while its spatial support directs a fixed share of future-RGB supervision toward regions of projected robot motion. KineWorld comprises two components. \textbf{Kinematic Transport Lifting
(KTL)} constructs camera-aligned transport and injects its latent representation
$z^u$ into the video generator as a clean structural condition.
\textbf{Transport-Aware World Diffusion (TAWD)} derives calibrated support $d$
from the transport fields and constructs normalized token weights for the
flow-matching objective. KTL supplies motion conditions during both training
and inference, whereas TAWD modifies only training-time supervision.

For a training clip starting at dense state index $s$, we sample $K$ visual keyframes
with temporal stride $\delta$ and define $L=(K-1)\delta$:
\begin{equation}
    \bar A=\{a_t\}_{t=s}^{s+L-1},\qquad
    \bar I_j=I_{s+j\delta},\qquad
    j=0,\ldots,K-1.
    \label{eq:clip_sampling}
\end{equation}
Starting from the initial robot configuration $q_s$, the commanded action
sequence is executed through the embodiment-specific controller,
$
    \hat q_0=q_s,
    \hat q_{\tau+1}=F_{\mathcal E}(\hat q_\tau,a_{s+\tau}),
    \tau=0,\ldots,L-1,
$
and sampled at the same temporal stride as
$
    \bar q_j=\hat q_{j\delta},
    j=0,\ldots,K-1.
$
The resulting rollout depends only on the initial robot configuration, the
commanded end-effector targets, and the embodiment-specific controller,
without access to future RGB observations. Given $\bar q_0$, $\bar A$, the
robot embodiment $\mathcal E$, and calibrated camera $\mathcal C$, Kinematic
Transport Lifting constructs
\begin{equation}
    U
    =
    \Gamma_{\mathcal E,\mathcal C}(\bar q_0,\bar A)
    =
    \{u_j\}_{j=0}^{K-1}.
    \label{eq:kineworld_transport}
\end{equation}
We use $K=9$ and $\delta=4$, corresponding to nine keyframes over 32 control
transitions.

\subsection{Kinematic Transport Lifting}
\label{sec:ktl}

\paragraph{Renderer-Derived Camera Transport.}
For every sampled robot configuration, a fixed renderer generates a
robot-only RGB frame and its visible robot mask:
\begin{equation}
    (R_j,\mathcal M_j)
    =
    \operatorname{Render}_{\mathcal E,\mathcal C}
    (\bar q_j),
    j=0,\ldots,K-1,
    \label{eq:robot_render}
\end{equation}
where $R_j\in\mathbb R^{H\times W\times 3}$ and
$\mathcal M_j\in\{0,1\}^{H\times W}$. We use a fixed background texture
during robot-only rendering to stabilize correspondence estimation.

For each pair of consecutive rendered states, RAFT estimates forward
displacement on the source image grid. The estimate is restricted to the
visible source-robot support:
\begin{equation}
    u_j
    =
    \begin{cases}
        0,
        & j=0,\\[1mm]
        \mathcal M_{j-1}
        \odot
        \operatorname{RAFT}(R_{j-1},R_j),
        & 1\leq j\leq K-1.
    \end{cases}
    \label{eq:robot_transport}
\end{equation}
The binary mask is broadcast over the two displacement channels. Each
$u_j:\Omega\rightarrow\mathbb R^2$ therefore defines a camera-aligned
transport field over image domain $\Omega$, where
$u_j(p)=(u_{j,x}(p),u_{j,y}(p))$.
RAFT~\citep{teed2020raft} estimates image-plane correspondences between consecutive robot-only renders, and the source mask restricts the resulting field to visible robot pixels. We retain this render-and-estimate construction, keeping the transport-conditioning pipeline fixed when studying supervision allocation. The resulting $u_j$ is therefore an image-based estimate of projected robot motion, rather than an analytic projection of surface trajectories.

\paragraph{Video-Native Transport Condition.}
For each pixel $p$, we define the transport magnitude and direction as
$m_j(p)=\lVert u_j(p)\rVert_2$ and
$\phi_j(p)=\operatorname{atan2}(u_{j,y}(p),u_{j,x}(p))$, respectively.
We encode direction as hue and clipped magnitude as saturation:
\begin{equation}
\begin{aligned}
    h_j(p)
    =
    \frac{\phi_j(p)+\pi}{2\pi},
    \eta_j(p)
    =
    \operatorname{clip}
    \left(
        \frac{m_j(p)}{M_{\mathrm{px}}},
        0,1
    \right),
    G_j(p)
    =
    \operatorname{HSV2RGB}
    \bigl(
        h_j(p),\eta_j(p),1
    \bigr).
\end{aligned}
\label{eq:transport_codec}
\end{equation}
Here, $M_{\mathrm{px}}>0$ is a fixed displacement-clipping threshold.
Zero displacement has zero saturation and is therefore mapped to white.

The RGB and transport clips are encoded separately:
\begin{equation}
z^x=\mathcal E_{\mathrm{VAE}}(\bar I_{0:K-1}),\quad
z^u=\mathcal E_{\mathrm{VAE}}(G_{0:K-1}),\quad
z^x,z^u\in\mathbb R^{C_z\times Q\times H_z\times W_z}.
\end{equation}
The encoded transport $z^u$ serves as a structural condition for RGB
generation. TAWD computes its supervision weights directly from the transport
fields $u_j$, rather than from their VAE latents. This separates supervision
allocation from possible compression artifacts in the transport representation,
although such artifacts may still affect the conditioning stream.

\paragraph{Clean Transport Conditioning.}
KTL supplies the transport-conditioning stream of the Wan2.2-based generator.
The RGB latent follows the flow-matching path defined in
Equation~\ref{eq:flow_path}, while the transport latent remains clean:
$\tilde z^u_{\tau,\ell}=z^u_\ell$ for
$\ell=0,\ldots,Q-1$. The model therefore receives a clean first RGB latent
frame, noised future RGB latents, and a clean transport sequence:
\begin{equation}
    \hat v_\theta
    =
    v_\theta
    \bigl(
        \tilde z^x_\tau,
        \tilde z^u_\tau,
        \tau
        \mid c
    \bigr).
    \label{eq:kineworld_flow_prediction}
\end{equation}
Here, \emph{clean transport} means that diffusion noise is not added to the
transport input. Its hidden representations can nevertheless interact with
the RGB stream throughout denoising.

RGB and transport latents are patchified separately, augmented with stream
identifiers, and assigned aligned spatiotemporal positions. Let $X_b^x$ and
$X_b^u$ denote the RGB and transport token sequences entering Transformer
block $b$, respectively. Their joint self-attention is written as
\begin{equation}
    H_b=[X_b^x;X_b^u],
    \Delta H_b=\operatorname{MSA}_b(H_b),
    [\Delta X_b^x;\Delta X_b^u]
    =
    \operatorname{Split}(\Delta H_b),
     b=1,\ldots,B.
    \label{eq:joint_attention}
\end{equation}
The two streams interact through joint self-attention throughout RGB denoising.
TAWD leaves this conditioning architecture unchanged and modifies only the
training objective. The computational cost of transport conditioning, including
joint attention, is accounted for separately in Appendix~\ref{app:efficiency_audit}.

\subsection{Transport-Aware World Diffusion}
\label{sec:tawd}

\paragraph{Latent Transport Support Calibration.}
Raw transport magnitude is not directly comparable across trajectories because
it varies with camera scale, action amplitude, visible robot area, and temporal
stride. We therefore construct a bounded activity map directly from the
source-masked transport:
\begin{equation}
    r_j(p)
    =
    \operatorname{clip}
    \left(
        \frac{\lVert u_j(p)\rVert_2}{M_{\mathrm{px}}},
        0,1
    \right),
    \qquad
    j=0,\ldots,K-1.
    \label{eq:raw_transport_support}
\end{equation}
Because $u_j$ has already been restricted by $\mathcal M_{j-1}$ in
Equation~\ref{eq:robot_transport}, the visibility mask is not applied again.
In particular, $r_0=0$ follows from $u_0=0$.

A fixed, non-learned operator $\mathcal P_{\mathrm{lat}}$ performs
spatiotemporal average pooling according to the video VAE compression ratios
and aligns the activity sequence with the RGB latent grid:
\begin{equation}
    d_{\ell hw}
    =
    \operatorname{sg}
    \left[
        \mathcal P_{\mathrm{lat}}
        (r_{0:K-1})_{\ell hw}
    \right],
    \qquad
    d_{0hw}=0.
    \label{eq:latent_transport_support}
\end{equation}
Here, $d\in[0,1]^{Q\times H_z\times W_z}$ and
$\operatorname{sg}(\cdot)$ denotes stop-gradient. The index $\ell$ denotes
latent time, whereas $j$ denotes an input keyframe. Computing $d$ from raw
transport rather than from the learned transport latent $z^u$ provides an
explicit zero-motion reference and prevents the transport encoder from
adapting the supervision weights.

\paragraph{Budget-Normalized Flow Matching.}
We reuse the future-token set $\mathcal V$ from
Equation~\ref{eq:future_token_set}. For a flattened future latent position
$i=(\ell,h,w)\in\mathcal V$, let $d_i$ denote the corresponding calibrated
transport support and define
$
    S_d
    =
    \sum_{i\in\mathcal V}d_i.
    \label{eq:total_support_mass}
$
The calibrated support is converted into a sample-wise transport distribution
with an explicit zero-motion fallback:
\begin{equation}
    \rho_i^{\mathrm{tr}}
    =
    \begin{cases}
        \displaystyle
        \frac{d_i}{S_d},
        & S_d>\varepsilon_{\mathrm{sup}},\\[3mm]
        \displaystyle
        \frac{1}{N},
        & S_d\leq\varepsilon_{\mathrm{sup}},
    \end{cases}
    \qquad
    \sum_{i\in\mathcal V}
    \rho_i^{\mathrm{tr}}
    =1,
    \label{eq:transport_distribution}
\end{equation}
where $\varepsilon_{\mathrm{sup}}>0$ is a small numerical threshold.

For a non-negative transport-strength parameter $\beta$, we define
$\alpha_\beta=\beta/(1+\beta)\in[0,1)$ and construct the per-token weights:
\begin{equation}
    w_i
    =
    N
    \left[
        (1-\alpha_\beta)\frac{1}{N}
        +
        \alpha_\beta\rho_i^{\mathrm{tr}}
    \right],
    \qquad
    \frac{1}{N}
    \sum_{i\in\mathcal V}w_i
    =
    1.
    \label{eq:transport_weight}
\end{equation}
The uniform component preserves scene-wide supervision, while the transport
component reallocates an $\alpha_\beta$ fraction of the aggregate token weight
according to projected robot-motion support. When $\beta=0$, $w_i=1$, exactly
recovering uniform flow matching. When no valid transport is present, the
fallback in Equation~\ref{eq:transport_distribution} also produces uniform
weights.

Using the token error $e_i(\theta)$ defined in
Equation~\ref{eq:token_error}, the KineWorld objective is
\begin{equation}
    \mathcal L_{\mathrm{Kine}}
    =
    \mathbb E_{\bar I_{0:K-1},\bar A,c,\tau,\epsilon}
    \left[
        \frac{1}{N}
        \sum_{i\in\mathcal V}
        \operatorname{sg}(w_i)
        e_i(\theta)
    \right].
    \label{eq:kine_loss}
\end{equation}
Because token weights are normalized to unit mean, transport magnitude and support only redistribute supervision. The calibrated map highlights projected robot-motion support rather than object motion or all downstream physical effects, while the uniform component preserves supervision over other scene regions.
\begin{table*}[!t]
\begingroup
\scriptsize
\centering
\addtolength{\tabcolsep}{0.1pt}
\caption{\textbf{Single-View Reference Profiles.} EWM denotes EWMScore-P \citep{shang2026worldarena2}; metric definitions and aggregation appear in Appendix~\ref{app:metric_aggregation}. These aggregates do not constitute a shared RoboTwin~2.0 evaluation. Source status is documented in Appendix~\ref{app:result_provenance}.}
\renewcommand\arraystretch{0.7}
\tabcolsep=2.1mm
\resizebox{\linewidth}{!}{
\begin{tabular}{l c ccc ccc}
\toprule
\multirow{2}{*}{$\operatorname{Method}$} & \multicolumn{1}{c}{$\operatorname{Overall}$} & \multicolumn{3}{c}{$\operatorname{Generation\ Quality}$} & \multicolumn{3}{c}{$\operatorname{Embodied\ Capability}$} \\
\cmidrule(lr){2-2}\cmidrule(lr){3-5}\cmidrule(lr){6-8}
& $\operatorname{EWM}\uparrow$ & $\operatorname{Vis.}\uparrow$ & $\operatorname{Mot.}\uparrow$ & $\operatorname{Cont.}\uparrow$ & $\operatorname{Phys.}\uparrow$ & $\operatorname{3D}\uparrow$ & $\operatorname{Ctrl.}\uparrow$ \\
\midrule
JF\_World \citep{dreamxphi2026} & 64.85 & 66.52 & 30.66 & 56.91 & 69.27 & 97.88 & 88.06 \\
BWM-Super \citep{bwm2026} & 64.30 & 67.20 & 30.16 & 58.36 & 64.15 & 97.35 & 87.19 \\
BWM-Turbo \citep{bwm2026} & 63.96 & 66.99 & 30.10 & 57.22 & 64.46 & 97.74 & 86.05 \\
FlowWAM-FiveAges \citep{chen2026flowwam} & 63.87 & 66.60 & 29.92 & 54.47 & 65.97 & 98.48 & 88.09 \\
WoVR\_Plus \citep{jiang2026wovr} & 57.89 & 58.37 & 30.52 & 57.91 & 47.95 & 85.31 & 80.75 \\
WorldScape-0.2 \citep{fang2026worldscapemoe} & 57.55 & 58.00 & 30.84 & 55.41 & 47.61 & 88.49 & 79.15 \\
PF-Cosmos \citep{zhang2026physisforcing} & 51.46 & 48.23 & 29.57 & 50.88 & 33.76 & 88.18 & 70.99 \\
Wan2.2 \citep{wanteam2025wan} & 49.21 & 49.55 & 26.89 & 51.51 & 29.24 & 81.31 & 66.57 \\
GigaWorld-0 \citep{gigaworldteam2025} & 48.06 & 46.09 & 30.37 & 46.49 & 31.35 & 76.10 & 68.57 \\
Vidar \citep{feng2025vidar} & 47.13 & 47.61 & 29.13 & 44.76 & 30.02 & 74.63 & 66.60 \\
IRASim \citep{zhu2024irasim} & 44.97 & 46.25 & 27.86 & 41.39 & 30.28 & 70.62 & 63.11 \\
\midrule
\rowcolor{red!5} KineWorld (Ours) & 68.95 & 67.09 & 33.36 & 69.01 & 76.49 & 98.53 & 87.94 \\
\bottomrule
\end{tabular}
}
\label{tab:robotwin_singleview_profiles}
\par
\endgroup
\vspace{0pt}
\begingroup
\scriptsize
\centering
\addtolength{\tabcolsep}{0.1pt}
\caption{\textbf{Single-View Local Diagnostics.} Reported video-quality measurements; metric scales and evaluation scope are detailed in Appendix~\ref{app:result_provenance}.}
\renewcommand\arraystretch{0.7}
\tabcolsep=2.2mm
\resizebox{\linewidth}{!}{
\begin{tabular}{l cccccccc}
\toprule
$\operatorname{Method}$ & $\operatorname{AES}\uparrow$ & $\operatorname{IMG}\uparrow$ & $\operatorname{BG}\uparrow$ & $\operatorname{SC}\uparrow$ & $\operatorname{PC}\uparrow$ & $\operatorname{INT}\uparrow$ & $\operatorname{PER}\uparrow$ & $\operatorname{IF}\uparrow$ \\
\midrule
Wan2.2-TI2V-5B \citep{wanteam2025wan} & 42.10 & 44.42 & 85.17 & \textbf{80.90} & 0.96 & 68.68 & 88.52 & 76.00 \\
\rowcolor{red!5} KineWorld & \textbf{45.26} & \textbf{61.21} & \textbf{87.19} & 75.53 & \textbf{16.43} & \textbf{79.20} & \textbf{97.30} & \textbf{85.84} \\
\bottomrule
\end{tabular}
}
\label{tab:robotwin_singleview_diagnostics}
\par
\endgroup
\vspace{0pt}
\begingroup
\scriptsize
\centering
\addtolength{\tabcolsep}{0.1pt}
\fontencoding{T1}\fontfamily{ptm}\selectfont
\caption{\textbf{Multi-View Reference Profiles.} TWB-Score is the TriWorldBench overall score \citep{triworldbench2026}; definitions appear in Appendix~\ref{app:twb_metrics}. Bold marks column maxima. These profiles are not a shared RoboTwin~2.0 evaluation. Components appear in Table~\ref{tab:robotwin_multiview_components}.}
\renewcommand\arraystretch{0.7}
\tabcolsep=2.1mm
\resizebox{\linewidth}{!}{
\begin{tabular}{l c ccc ccc}
\toprule
\multirow{2}{*}{$\operatorname{Method}$} & \multicolumn{1}{c}{$\operatorname{Overall}$} & \multicolumn{3}{c}{$\operatorname{Embodied\ Fidelity}$} & \multicolumn{3}{c}{$\operatorname{Generation\ Quality}$} \\
\cmidrule(lr){2-2}\cmidrule(lr){3-5}\cmidrule(lr){6-8}
& $\operatorname{TWB\mbox{-}Score}\uparrow$ & $\operatorname{Tri.}\uparrow$ & $\operatorname{Task}\uparrow$ & $\operatorname{Phys.}\uparrow$ & $\operatorname{Mot.}\uparrow$ & $\operatorname{Temp.}\uparrow$ & $\operatorname{Vis.}\uparrow$ \\
\midrule
BWM \citep{bwm2026} & \textbf{65.54} & 81.87 & \textbf{86.05} & 60.40 & 41.29 & 62.81 & \textbf{31.42} \\
WoVR\_Plus \citep{jiang2026wovr} & 65.39 & \textbf{82.37} & 82.80 & 57.12 & \textbf{43.68} & 64.10 & 31.14 \\
DreamDojo \citep{gao2026dreamdojo} & 51.72 & 69.63 & 56.24 & 43.84 & 27.96 & 60.84 & 21.02 \\
Motus \citep{bi2025motus} & 42.35 & 66.70 & 49.69 & 34.60 & 24.63 & 26.56 & 16.26 \\
Genie Envisioner \citep{liao2025genieenvisioner} & 40.73 & 62.39 & 33.17 & 54.00 & 20.17 & 32.18 & 17.46 \\
Ctrl-World \citep{guo2025ctrlworld} & 42.35 & 57.42 & 43.72 & 34.67 & 29.28 & 46.17 & 16.64 \\
\rowcolor{red!5} KineWorld (Ours) & 54.82 & 66.36 & 50.72 & \textbf{63.24} & 32.28 & \textbf{69.90} & 29.15 \\
\bottomrule
\end{tabular}
}
\label{tab:robotwin_multiview_profiles}
\par
\endgroup
\vspace{0pt}
\begingroup
\scriptsize
\centering
\addtolength{\tabcolsep}{0.1pt}
\caption{\textbf{Core Module Ablation.} EWM denotes EWMScore-P. NFE counts denoiser evaluations.}
\renewcommand\arraystretch{0.7}
\tabcolsep=0.5mm
\resizebox{\linewidth}{!}{
\begin{tabular}{cc ccccccc cc ccc}
\toprule
\multicolumn{2}{c}{$\operatorname{Core\ Module}$} & \multicolumn{7}{c}{$\operatorname{Retained\ Video\ Dimensions}$} & \multicolumn{2}{c}{$\operatorname{Action\ Response\ (\%)}$} & \multicolumn{3}{c}{$\operatorname{Efficiency}$} \\
\cmidrule(lr){1-2}\cmidrule(lr){3-9}\cmidrule(lr){10-11}\cmidrule(lr){12-14}
$\operatorname{KTL}$ & $\operatorname{TAWD}$ & $\operatorname{EWM}\uparrow$ & $\operatorname{Visual}\uparrow$ & $\operatorname{Motion}\uparrow$ & $\operatorname{Content}\uparrow$ & $\operatorname{Physics}\uparrow$ & $\operatorname{3D}\uparrow$ & $\operatorname{Control}\uparrow$ & $\operatorname{INT}\uparrow$ & $\operatorname{TA}\uparrow$ & $\operatorname{Params\ (B)}\downarrow$ & $\operatorname{NFE}\downarrow$ & $\operatorname{FPS}\uparrow$ \\
\midrule
-- & -- & 49.21 & 49.55 & 26.89 & 51.51 & 29.24 & 81.31 & 66.57 & 44.32 & 14.15 & 5.000 & 50 & 2.80 \\
\midrule
\checkmark & -- & 65.99 & 64.46 & 32.39 & 66.39 & 69.40 & 95.95 & 84.73 & 76.69 & 62.11 & 5.001 & 25 & 3.01 \\
\rowcolor{red!5} \checkmark & \checkmark & 68.95 & 67.09 & 33.36 & 69.01 & 76.49 & 98.53 & 87.94 & 82.40 & 70.57 & 5.001 & 25 & 3.02 \\
\bottomrule
\end{tabular}
}
\label{tab:core_module_ablation}
\par
\endgroup
\end{table*}
\begin{table*}[!t]
\begingroup
\scriptsize
\centering
\addtolength{\tabcolsep}{0.1pt}
\caption{\textbf{KTL Ablation.} Three-view validation diagnostic on 100 episodes (300 videos per variant). Dataset scope and metric definitions appear in Appendix~\ref{app:ktl_transport_protocol}.}
\renewcommand\arraystretch{0.7}
\tabcolsep=0.1mm
\resizebox{\linewidth}{!}{
\begin{tabular}{l cccc ccc cc}
\toprule
\multirow{2}{*}{$\operatorname{Method}$} & \multicolumn{4}{c}{$\operatorname{Overall\ Fidelity}$} & \multicolumn{3}{c}{$\operatorname{View\ PSNR\ (dB)}$} & \multicolumn{2}{c}{$\operatorname{Boundary\ PSNR\ (dB)}$} \\
\cmidrule(lr){2-5}\cmidrule(lr){6-8}\cmidrule(lr){9-10}
& $\operatorname{PSNR}\uparrow$ & $\operatorname{SSIM}\uparrow$ & $\operatorname{N\mbox{-}MAE}\downarrow$ & $\operatorname{Mot.\ Corr.}\uparrow$ & $\operatorname{Head}\uparrow$ & $\operatorname{Left}\uparrow$ & $\operatorname{Right}\uparrow$ & $\operatorname{First}\uparrow$ & $\operatorname{Last}\uparrow$ \\
\midrule
KineWorld (white wrist flow) & 35.35 & 0.9622 & 16.532 & 0.9132 & 36.10 & 36.13 & 33.82 & 42.01 & 32.32 \\
\rowcolor{red!5} KineWorld (all-view RAFT) & \textbf{42.76} & \textbf{0.9860} & \textbf{5.713} & \textbf{0.9953} & \textbf{41.88} & \textbf{43.28} & \textbf{43.13} & \textbf{42.65} & \textbf{42.76} \\
\bottomrule
\end{tabular}
}
\label{tab:ktl_design_ablation}
\par
\endgroup
\vspace{0pt}
\begingroup
\scriptsize
\centering
\addtolength{\tabcolsep}{0.1pt}
\caption{\textbf{TAWD Ablation.} EWM denotes EWMScore-P. Both variants start from the same step-315 checkpoint and are trained for 50 updates with the same data order, seeds, KTL condition, and inference and evaluation settings; only $\beta$ changes. Intervals use 20,000 episode-paired bootstrap resamples over $M=1000$ complete episodes (Appendix~\ref{app:statistical_protocol}).}
\renewcommand\arraystretch{0.7}
\tabcolsep=1.4mm
\resizebox{\linewidth}{!}{
\begin{tabular}{l c ccccc}
\toprule
\multirow{2}{*}{$\operatorname{Matched\ Continuation}$} & \multirow{2}{*}{$\beta$} & \multicolumn{5}{c}{$\operatorname{Video\ Metrics}$} \\
\cmidrule(lr){3-7}
& & $\operatorname{EWM}\uparrow$ & $\operatorname{Motion}\uparrow$ & $\operatorname{TA}\uparrow$ & $\operatorname{INT}\uparrow$ & $\operatorname{Visual}\uparrow$ \\
\midrule
Uniform weighting & 0 & 66.10 & 32.50 & 62.80 & 77.00 & 64.80 \\
\rowcolor{red!5} TAWD & 1 & \textbf{67.85} & \textbf{33.35} & \textbf{68.40} & \textbf{81.20} & \textbf{65.40} \\
$\Delta$ (TAWD $-$ Uniform) & -- & \textbf{$+1.75$} & \textbf{$+0.85$} & \textbf{$+5.60$} & \textbf{$+4.20$} & $+0.60$ \\
95\% paired CI & -- & $[+1.10,+2.37]$ & $[+0.25,+1.39]$ & $[+3.20,+7.90]$ & $[+2.10,+6.00]$ & $[-0.10,+1.20]$ \\
\bottomrule
\end{tabular}
}
\label{tab:tawd_design_ablation}
\par
\endgroup
\vspace{0pt}
\begingroup
\scriptsize
\centering
\addtolength{\tabcolsep}{0.1pt}
\caption{\textbf{Model Efficiency and Inference Speed.} Comparison of model size and inference efficiency across representative world models. KineWorld uses precomputed kinematic transport for the reported inference speed.}
\color{black}
\renewcommand\arraystretch{0.7}
\tabcolsep=0.5mm
\resizebox{\linewidth}{!}{
\begin{tabular}{l l cc c c}
\toprule
\multirow{2}{*}{$\operatorname{Method}$} & \multirow{2}{*}{$\operatorname{Type}$} & \multicolumn{2}{c}{$\operatorname{Efficiency}$} & \multirow{2}{*}{$\operatorname{Output}$} & \multirow{2}{*}{$\operatorname{Hardware}$} \\
\cmidrule(lr){3-4}
& & $\operatorname{Params\ (B)}\downarrow$ & $\operatorname{FPS}\uparrow$ & & \\
\midrule
IRASim-XL \citep{zhu2024irasim} & Robot WAM & 0.679 & 0.53 &$288\times512$ & 1$\times$A100 \\
Wan2.2-TI2V-5B \citep{wanteam2025wan,zhao2026rynnworld} & General I2V & 5.000 & 2.80 & $480\times832$  & 1$\times$H100 \\
RynnWorld-Teleop-Causal \citep{zhao2026rynnworld} & Robot WAM & 5.000 & 40.00 & $480\times832$ & 1$\times$H100 \\
GE-Sim 2.0 \citep{qiu2026gesim2} & Robot WAM & 2.000 & 10.87 & -- & 1$\times$H100 \\
\rowcolor{red!5} KineWorld (Ours) & Robot WAM & 5.001 & 3.02 & $640\times480$ & 1 HCU \\
\bottomrule
\end{tabular}
}
\label{tab:model_efficiency_comparison}
\par
\endgroup
\end{table*}

\paragraph{Auxiliary Action Objective.}
We retain the inherited action head for checkpoint-matched comparisons. Let
$\theta_{\mathrm{wm}}$ and $\theta_{\mathrm{act}}$ denote the disjoint
parameter sets of the world generator and action head, respectively. The
action head operates on stopped world-model features:
\begin{equation}
    \hat A
    =
    g_{\theta_{\mathrm{act}}}
    \left(
        \operatorname{sg}
        \bigl(
            H_{\theta_{\mathrm{wm}}}
        \bigr)
    \right).
    \label{eq:auxiliary_action_prediction}
\end{equation}
The complete training objective is
\begin{equation}
    \mathcal L
    =
    \mathcal L_{\mathrm{Kine}}
    (\theta_{\mathrm{wm}})
    +
    \lambda_{\mathrm{act}}
    \mathcal L_{\mathrm{act}}
    \left(
        \theta_{\mathrm{act}};
        \operatorname{sg}
        \bigl(
            H_{\theta_{\mathrm{wm}}}
        \bigr)
    \right).
    \label{eq:total_loss}
\end{equation}
Because the action head receives stopped features and uses a disjoint
parameter set,
$\nabla_{\theta_{\mathrm{wm}}}\mathcal L_{\mathrm{act}}=0$.
The auxiliary branch is unchanged across all checkpoint-matched comparisons
and is not part of the KineWorld-specific optimization.

\paragraph{Training and Inference.}
During training, TAWD derives the calibrated support $d$ from the
renderer-derived transport and converts it into $\rho^{\mathrm{tr}}$ and $w$
for reweighting the flow-matching objective. At inference time, the support
calibration and loss-weighting branch is not required. The clean transport
latent $z^u$ remains as a structural condition, while the learned vector field
is integrated from $\tau=1$ to $\tau=0$ to obtain future RGB latents, which
are decoded into future video frames.

\section{Experiments}

\subsection{Experimental Setup}
\label{sec:experimental_setup}

\paragraph{Dataset and Task Definition.}
We consider action-conditioned video prediction using RoboTwin~2.0 Aloha--AgileX Clean-50 demonstrations \citep{chen2025robotwin2}. The collection contains 1,800 development-training, 200 validation, and 500 held-out episodes, the combined training set contains the first two partitions. Single-view comparisons, multi-view comparisons, and local generation diagnostics are reported separately.

\paragraph{Implementation Details.}
We initialize the video generator from Wan2.2-TI2V-5B \citep{wanteam2025wan}. KTL estimates camera-aligned transport with torchvision RAFT \citep{teed2020raft} and encodes it through the video VAE, while TAWD reweights only the future-RGB objective without adding sampling steps. Appendix~\ref{app:implementation} describes the training procedure and controlled-comparison settings; Appendix~\ref{app:efficiency_audit} provides parameter and runtime accounting.

\subsection{Reference Profiles and Local Diagnostics}
\label{sec:retained_results}
\paragraph{Single-View Result Profiles.}
Table~\ref{tab:robotwin_singleview_profiles} summarizes reported single-view system profiles; Table~\ref{tab:full_metrics} provides all 15 component scores. The KineWorld candidate profile has an EWMScore-P of 68.95 and Trajectory Accuracy of 70.57. These profiles provide system-level context rather than a common-training comparison on the Clean-50 held-out set.

\paragraph{Reported Local Diagnostics.}
Table~\ref{tab:robotwin_singleview_diagnostics} reports the local diagnostic profiles for Wan2.2-TI2V-5B and KineWorld. KineWorld has higher reported Image Quality (61.21 vs. 44.42), Interaction Quality (79.20 vs. 68.68), and Instruction Following (85.84 vs. 76.00), but lower Subject Consistency (75.53 vs. 80.90). These measurements are analyzed separately from the aggregate profile in Table~\ref{tab:robotwin_singleview_profiles}, preserving their original evaluation scope and score scales.

\paragraph{Multi-View Result Profiles.}
Table~\ref{tab:robotwin_multiview_profiles} reports a TWB-Score of 54.82 for KineWorld, placing it third among the seven listed profiles. KineWorld has the highest Physical and 3D Coherence (63.24) and Temporal Consistency (69.90), exceeding the next-best reported values by 2.84 and 5.80 points, respectively. Tri-View Consistency (66.36) and Task Alignment (50.72) remain below the leading entries, leaving room to improve cross-view and task fidelity.

\subsection{Ablation Study}
\label{sec:ablations}

\paragraph{Core Module Ablation.}
Table \ref{tab:core_module_ablation} shows that KTL alone greatly improves all metrics over the baseline. EWM rises from 49.21 to 65.99, Physics from 29.24 to 69.40, and TA from 14.15 to 62.11, while NFE drops from 50 to 25. Adding TAWD further boosts EWM to 68.95, Physics to 76.49, INT to 82.40, and TA to 70.57, with almost unchanged parameters and slightly higher FPS. These results demonstrate that KTL and TAWD are complementary core modules that deliver substantial gains in action response, physical plausibility, and generation efficiency at negligible overhead.

\paragraph{KTL Design.}
Table~\ref{tab:ktl_design_ablation} reports a fixed-checkpoint diagnostic on 100 three-view episodes (300 videos per variant). All-view RAFT guidance improves all metrics over white wrist-flow placeholders, raising PSNR from 35.35 to 42.76 dB, SSIM from 0.9622 to 0.9860, and motion correlation from 0.9132 to 0.9953, while reducing N-MAE from 16.532 to 5.713. Gains are consistent across views and boundary frames. These results validate KTL as effective camera-aligned transport for motion conditioning.

\begin{figure}[t]
    \centering
    \includegraphics[width=\linewidth]{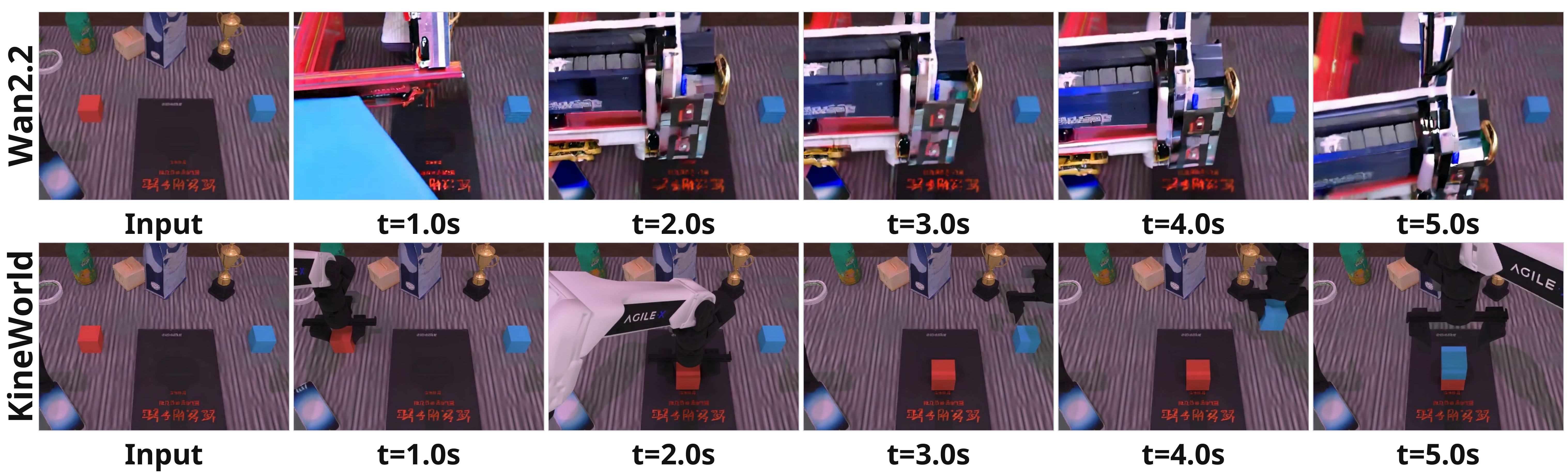}
    \caption{\textbf{Retained qualitative comparison between Wan2.2 and KineWorld.}
    Decoded episode-550 frames at $t=1.0$, $2.0$, $3.0$, $4.0$, and $5.0$~s from 24-fps videos. This is a descriptive rollout comparison, not an action- or checkpoint-matched ablation.}
    \label{fig:kineworld_wan_visual}
\end{figure}

\paragraph{TAWD Design.}
Table~\ref{tab:tawd_design_ablation} compares uniform CFM ($\beta=0$) with TAWD ($\beta=1$) under matched 50-update continuations from the same step-315 checkpoint on 1,000 paired episodes, sharing all settings except the spatial allocation of a fixed token weight. TAWD improves EWMScore-P by 1.75, Trajectory Accuracy by 5.60, Interaction Quality by 4.20, and Motion Quality by 0.85, while Visual Quality remains inconclusive. These results validate TAWD as an effective mechanism for reallocating future-RGB supervision toward action-responsive regions.

\paragraph{Model Size and Inference Speed.}
Table~\ref{tab:model_efficiency_comparison} reports model size and speed. KineWorld uses 5.001B parameters (KTL adds 1.19M, TAWD none) and reaches 3.02 FPS per HCU worker with precomputed transport. Baselines are literature-reported under method-specific settings, not local speed tests, and are not directly comparable across hardware and inference settings.
\subsection{Visualization}
\label{sec:qualitative}
Figure~\ref{fig:kineworld_wan_visual} qualitatively compares Wan2.2 and KineWorld rollouts on episode 550, showing decoded frames at $t=1.0-5.0$~s from 24-fps videos. KineWorld preserves morphology and layout and yields action-responsive, temporally consistent rollouts, advancing action-consequence modeling.


\section{Conclusion}
\label{sec:conclusion}

We presented KineWorld, a kinematics-guided framework for action-conditioned video prediction. KTL constructs camera-aligned transport, and TAWD calibrates its support and uses it to redistribute future-RGB supervision between action-responsive regions and the full scene. The objective preserves the total token weight and recovers uniform supervision under zero motion, without adding inference parameters or denoising steps. Trained on ALOHA--AgileX bimanual manipulation data from \textbf{RoboTwin~2.0}, KineWorld achieves an \textbf{EWMScore-P of 68.95} in single-view evaluation and a \textbf{TWB-Score of 54.82} in multi-view evaluation. These results highlight the potential of kinematics-guided supervision to advance world models from appearance fitting toward action-consequence modeling for embodied decision-making.

\paragraph{Limitation and Future Work.}
KineWorld relies on calibrated robot and camera models, and its transport support does not explicitly capture object motion or contact dynamics. Future work will investigate calibration-robust transport estimation, object-aware supervision, and contact-aware world modeling.

\section*{AI Use Disclosure}

Generative AI tools assisted with research framing, manuscript editing, code inspection, experimental-protocol design, literature discovery, and reference and figure formatting. AI-assisted editing did not generate experimental measurements. The authors are responsible for the scientific claims, numerical results, citations, and final manuscript.

\section*{Reproducibility Statement}
We provide a complete specification of the data splits, training and inference procedures, evaluation protocols, and statistical analysis to support reproduction of KineWorld. The matched comparison settings and objective definitions are described in the main text. We commit to publicly releasing the code, checkpoints, and data manifests upon acceptance.

\bibliographystyle{iclr2027_conference}
\bibliography{iclr2027_conference}

\documentclass{article}

\usepackage{iclr2027_conference,times}

\usepackage{amsmath}
\usepackage{amssymb}
\usepackage{array}
\usepackage{tabularx}
\usepackage{booktabs}
\usepackage{graphicx}
\usepackage{float}
\usepackage{placeins}
\usepackage{multirow}
\usepackage{xcolor}
\usepackage{colortbl}
\usepackage{xspace}
\usepackage{xr}
\usepackage{hyperref}
\hypersetup{hidelinks}
\usepackage{url}
\externaldocument[V3main-]{iclr2027_conferenceV3_main}
\newif\ifVThreeAppendixStandalone
\VThreeAppendixStandalonetrue

\newcommand{\TAWD}{\textsc{TAWD}\xspace}
\DeclareRobustCommand{\reviewold}[1]{\textcolor{blue}{#1}}
\DeclareRobustCommand{\reviewedit}[1]{\textcolor{red}{#1}}
\definecolor{kineworldTitleRed}{HTML}{F5222D}
\definecolor{kineworldTitleOrange}{HTML}{FA8C16}
\definecolor{kineworldTitleYellow}{HTML}{E8BE24}
\definecolor{kineworldTitleGreen}{HTML}{00A651}
\definecolor{kineworldTitleBlue}{HTML}{0072BC}
\definecolor{kineworldTitleCyan}{HTML}{00A6D6}
\definecolor{kineworldTitlePurple}{HTML}{9C44AA}
\definecolor{kineworldblue}{HTML}{2F5D7C}
\DeclareRobustCommand{\KineWorldColor}{\textcolor{kineworldblue}{KineWorld}}
\DeclareRobustCommand{\KineWorldTitleColor}{%
  \textcolor{kineworldTitleRed}{K}%
  \textcolor{kineworldTitleOrange}{i}%
  \textcolor{kineworldTitleYellow}{n}%
  \textcolor{kineworldTitleGreen}{e}%
  \textcolor{kineworldTitleBlue}{W}%
  \textcolor{kineworldTitleCyan}{o}%
  \textcolor{kineworldTitlePurple}{rld}%
}
\pdfstringdefDisableCommands{\def\KineWorldColor{KineWorld}\def\KineWorldTitleColor{KineWorld}\def\reviewold#1{#1}\def\reviewedit#1{#1}}
\title{\KineWorldTitleColor: Action-Induced Transport Fields\\for Embodied World Modeling\\Supplementary Material}
\author{}


\begin{document}
\ifVThreeAppendixStandalone
    \maketitle
    \setcounter{equation}{\getrefnumber{V3main-eq:total_loss}}
\else
    \clearpage
\fi
\newcommand{\appmainref}[1]{%
    \ifVThreeAppendixStandalone\ref{V3main-#1}\else\ref{#1}\fi}
\appendix
\section{Appendix}
\label{app:appendix}
\setcounter{figure}{0}
\renewcommand{\thefigure}{A.\arabic{figure}}
\renewcommand{\theHfigure}{appendix.A.\arabic{figure}}
\setcounter{table}{0}
\renewcommand{\thetable}{A.\arabic{table}}
\renewcommand{\theHtable}{appendix.A.\arabic{table}}

\newcommand{\appendixtablestyle}{%
    \centering
    \fontsize{6.66}{8}\selectfont
    \renewcommand{\arraystretch}{0.7}%
    \setlength{\tabcolsep}{2.1mm}%
    \setlength{\heavyrulewidth}{0.76pt}%
    \setlength{\lightrulewidth}{0.475pt}%
    \setlength{\cmidrulewidth}{0.285pt}%
}
\newcolumntype{Y}{>{\centering\arraybackslash}X}
\newenvironment{appendixtabular}[1]
    {\tabularx{\linewidth}{#1}}
    {\endtabularx}

This supplementary material provides additional descriptions and evaluations
of the proposed \textbf{KineWorld} framework. It is organized as follows.
\begin{itemize}
    \item \textbf{Appendix~\ref{app:broader_impacts}} discusses the
    \textbf{broader impacts}.
    \item \textbf{Appendix~\ref{app:contributions}} summarizes the
    \textbf{main contributions}.
    \item \textbf{Appendix~\ref{app:full_results}} describes the evaluated
    \textbf{RoboTwin~2.0 data, local protocols, and retained result profiles}.
    \item \textbf{Appendix~\ref{app:objective_derivation}} provides
    \textbf{additional details of KineWorld}.
    \item \textbf{Appendix~\ref{app:implementation}} presents the
    \textbf{implementation and optimization settings}.
    \item \textbf{Appendix~\ref{app:efficiency_audit}} reports the
    \textbf{audited model size and runtime}.
    \item \textbf{Appendix~\ref{app:statistical_protocol}} defines the
    \textbf{evaluation and statistical protocols}.
    \item \textbf{Appendix~\ref{app:additional_ablations}} reports
    \textbf{additional ablation results}.
    \item \textbf{Appendix~\ref{app:qualitative_protocol}} provides
    \textbf{additional qualitative results}.
    \item \textbf{Appendix~\ref{app:limitations}} discusses the
    \textbf{limitations and reproducibility requirements}.
\end{itemize}

\subsection{Broader Impacts}
\label{app:broader_impacts}
KineWorld may improve the action responsiveness of embodied world models by directing learning toward image regions that are causally associated with commanded robot motion. More faithful action-conditioned rollouts could support simulation, policy evaluation, and data generation while reducing unnecessary physical interaction. At the same time, generated videos remain predictions rather than certified physical trajectories: errors in calibration, contact modeling, or long-horizon generation can produce plausible but unsafe outcomes. KineWorld should therefore be used together with independent feasibility checks, collision detection, controller limits, and validation on the target robot.

\subsection{Contributions}
\label{app:contributions}
\paragraph{Problem Discovery.}
We distinguish action conditioning from supervision allocation. Action-induced motion can occupy only a small fraction of video-latent tokens, so uniform averaging of the RGB-generation objective can underemphasize prediction errors in action-responsive regions. Explicit motion conditions alone do not correct this weighting imbalance.

\paragraph{KineWorld Framework.}
We introduce KineWorld, a transport-aware embodied world-modeling framework that promotes robot kinematics from an auxiliary condition to a signal for allocating generative supervision. Given the initial embodiment configuration and commanded action sequence, KineWorld constructs camera-aligned transport support and redistributes token-wise supervision between global scene modeling and action-responsive regions while preserving the total token weight.

\paragraph{Core Modules.}
Kinematic Transport Lifting (KTL) rolls out the commanded configurations, renders the embodiment from the observation camera, and estimates source-supported inter-frame transport. Transport-Aware World Diffusion (TAWD) calibrates this activity on the video-latent grid, converts the calibrated support into a normalized mixture distribution, and uses mean-one weights to reallocate the conditional flow-matching objective, while preserving uniform supervision under zero motion and introducing no additional denoising steps.

\paragraph{Experiments.}
Our evaluation distinguishes retained system profiles, local generation diagnostics, and an executed matched comparison. In the matched comparison, two 50-update continuations start from the same step-315 checkpoint, keep KTL conditioning fixed, and differ only in the future-RGB weighting parameter $\beta$. Changing $\beta=0$ to $\beta=1$ improves EWMScore-P from 66.10 to 67.85 ($\Delta=+1.75$), with additional gains of 5.60 points in Trajectory Accuracy and 4.20 points in Interaction Quality.

\subsection{RoboTwin~2.0 Data, Local Protocols, and Retained Profiles}
\label{app:full_results}
\label{app:dataset_protocol}

\paragraph{Single-View Local Protocol.}
A single-view example comprises an initial head-camera observation, language instruction, initial robot configuration, and action trajectory. Future RGB frames serve as references for reference-based metrics. Following the action definition in Section~\appmainref{sec:embodied_world_modeling}, each arm is controlled by an absolute 6-DoF end-effector pose target and one gripper command, yielding a 14-D bimanual action vector. The reported $640\times480$, 24-fps videos use a pipeline-specific export convention. Resolution and playback rate do not specify the dataset split or the physical control frequency.

\paragraph{Data Partition.}
\label{app:data_split_mapping}
Clean-50 contains 2,500 episodes across 50 tasks. Within each task, episodes 0--35, 36--39, and 40--49 define the development-training, validation, and held-out partitions, respectively (Table~\ref{tab:data_split_mapping}). These partitions are episode-disjoint but share the same tasks. The combined 2,000-episode training set includes development training and validation; the 200-episode validation partition is therefore not unseen for a model trained on this combined set. Held-out denotes exclusion from training, not the number of completed model evaluations.

\begin{table*}[t]
\appendixtablestyle
\caption{\textbf{RoboTwin~2.0 Clean-50 Data Splits.} Indices are local to each of the 50 tasks. Counts refer to unique episodes; combined training includes the validation partition.}
\label{tab:data_split_mapping}
\begin{appendixtabular}{l Y Y Y}
\toprule
Partition & Episode indices per task & Episodes & Membership \\
\midrule
Development training & 0--35 & 1,800 & Disjoint from validation \\
Validation & 36--39 & 200 & Included in combined training \\
Combined training & 0--39 & 2,000 & Development $\cup$ validation \\
Held-out & 40--49 & 500 & Disjoint from both training sets \\
\bottomrule
\end{appendixtabular}
\end{table*}

\paragraph{Evaluation Collections and Result Tables.}
Single-view reference profiles appear in Tables~\appmainref{tab:robotwin_singleview_profiles} and~\ref{tab:full_metrics}, local single-view diagnostics in Tables~\appmainref{tab:robotwin_singleview_diagnostics} and~\ref{tab:local_singleview_full1000}, and multi-view reference profiles in Table~\appmainref{tab:robotwin_multiview_profiles}. The KTL diagnostic uses episodes 1--100 of a separately released three-view validation collection, with 300 videos per variant. The interpolation diagnostic in Table~\ref{tab:main_interpolation_ablation} uses episodes 17--32 of the single-view generation collection: 16 videos and 3,770 frames at $640\times480$ and 24 fps. These collection-local identifiers are not Clean-50 task--episode identifiers; reference profiles and diagnostics are not reassigned to the Clean-50 held-out partition.
Table~\appmainref{tab:tawd_design_ablation} reports the executed matched TAWD comparison over $M=1000$ paired single-view episodes. This comparison is interpreted separately from the retained system profiles and local diagnostic collections.

\paragraph{Counting and Filtering.}
An episode is counted once regardless of camera views or temporal clips. All 2,500 Clean-50 episodes passed data-preparation checks for required observations, videos, instructions, trajectories, and camera calibration; missing inputs trigger a failure rather than silent filtering. Screening against the separate 1,000-episode single-view test collection found no exact initial-image or action-array matches, so it excluded no additional episodes. The executed matched TAWD comparison uses this separate 1,000-episode single-view test collection; all $M=1000$ episode pairs are retained for the paired analysis. This screen tests exact duplication, not task-level or semantic independence.

\paragraph{Training Scope.}
The FlowWAM checkpoint used in the interpolation diagnostic was trained on the combined 2,000-episode Clean-50 set. This training-set assignment is specific to that checkpoint, rather than a shared setting for all reference profiles. External method scores retain their original training conditions.

\paragraph{Three-View Local Protocol.}
The three-view protocol uses synchronized head, left-wrist, and right-wrist observations together with an initial robot state, instruction, and action sequence. Each view uses its own camera calibration and transport condition, including time-dependent extrinsics for moving wrist cameras. The reported inference convention generates 17 uniformly spaced keyframes and expands them to the episode length. This convention describes temporal sampling and does not define episode membership. Reference-based metrics use the corresponding future observations from each camera.

\paragraph{Scope of Reported Results.}
\label{app:result_provenance}
The appendix separates four evidence layers: externally reported system profiles, the retained KineWorld candidate profile, local generation diagnostics, and the executed matched TAWD comparison in Table~\appmainref{tab:tawd_design_ablation}. The first three layers retain their original evaluation scope and do not provide common-training causal estimates. The matched comparison fixes initialization, training budget, KTL conditioning, data order, and evaluation, and changes only $\beta$. The retained KineWorld profile has EWMScore-P 68.95, whereas the matched $\beta=1$ continuation in Table~\appmainref{tab:tawd_design_ablation} has EWMScore-P 67.85. These values correspond to different training scopes and are not intended as two measurements of the same run. EWMScore-P follows the 15-component perceptual scoring convention defined in Appendix~\ref{app:metric_aggregation}; TWB-Score follows TriWorldBench's separate 19-component multi-view convention in Appendix~\ref{app:twb_metrics}. Neither is a newly proposed metric or a RoboTwin~2.0 task success rate. Method citations identify the underlying approaches; they do not establish common evaluation conditions. The objective-label interpretation is specified separately in Appendix~\ref{app:tawd_design_protocol}.

\begin{table*}[t]
\appendixtablestyle
\caption{\textbf{Complete Single-View Reference Profiles.} $\operatorname{EWM}$: EWMScore-P; $\operatorname{AES}$: Aesthetic Quality; $\operatorname{IMG}$: Image Quality; $\operatorname{JEPA}$: JEPA Similarity; $\operatorname{DYN}$: Dynamic Degree; $\operatorname{FL}$: Flow Score; $\operatorname{MS}$: Motion Smoothness; $\operatorname{BG}$: Background Consistency; $\operatorname{PC}$: Photometric Consistency; $\operatorname{SC}$: Subject Consistency; $\operatorname{INT}$: Interaction Quality; $\operatorname{TA}$: Trajectory Accuracy; $\operatorname{DEP}$: Depth Accuracy; $\operatorname{PER}$: Perspectivity; $\operatorname{IF}$: Instruction Following; $\operatorname{SA}$: Semantic Alignment. Original score scales are preserved. $^\dagger$ denotes the author-supplied candidate profile.}
\setlength{\tabcolsep}{0.25mm}
\renewcommand{\arraystretch}{1.0}
\resizebox{\linewidth}{!}{%
\begin{tabular}{l c ccc ccc ccc cc cc cc}
\toprule
\multirow{2}{*}{$\operatorname{Method}$} & \multicolumn{1}{c}{$\operatorname{Overall}$} & \multicolumn{3}{c}{$\operatorname{Visual}$} & \multicolumn{3}{c}{$\operatorname{Motion}$} & \multicolumn{3}{c}{$\operatorname{Content}$} & \multicolumn{2}{c}{$\operatorname{Physics}$} & \multicolumn{2}{c}{$\operatorname{3D}$} & \multicolumn{2}{c}{$\operatorname{Control}$} \\
\cmidrule(lr){2-2}\cmidrule(lr){3-5}\cmidrule(lr){6-8}\cmidrule(lr){9-11}\cmidrule(lr){12-13}\cmidrule(lr){14-15}\cmidrule(lr){16-17}
& EWM & AES & IMG & JEPA & DYN & FL & MS & BG & PC & SC & INT & TA & DEP & PER & IF & SA \\
\midrule
JF\_World \citep{dreamxphi2026} & 64.85 & 43.38 & 63.25 & 92.93 & 22.90 & 5.81 & 63.26 & 84.52 & 14.29 & 71.93 & 81.40 & 57.15 & 98.55 & 97.20 & 85.60 & 90.53 \\
BWM-Super \citep{bwm2026} & 64.30 & 44.39 & 62.38 & 94.83 & 22.01 & 5.77 & 62.70 & 84.31 & 18.92 & 71.86 & 79.20 & 49.09 & 97.51 & 97.20 & 85.20 & 89.19 \\
BWM-Turbo \citep{bwm2026} & 63.96 & 44.69 & 60.27 & 96.00 & 22.23 & 5.76 & 62.31 & 82.97 & 17.33 & 71.37 & 78.60 & 50.32 & 98.48 & 97.00 & 82.80 & 89.31 \\
FlowWAM-FiveAges \citep{chen2026flowwam} & 63.87 & 44.00 & 64.68 & 91.12 & 22.26 & 5.80 &
61.69 & 79.62 & 15.08 & 68.71 & 79.60 & 52.33 & 99.15 & 97.80 & 87.20 & 88.98 \\

WoVR Plus \citep{jiang2026wovr} & 57.89 & 40.34 & 46.78 & 87.99 & 22.92 & 5.80 &
62.83 & 86.99 & 15.46 & 71.27 & 67.20 & 28.69 & 85.02 & 85.60 & 73.00 & 88.50 \\

WorldScape-0.2 \citep{fang2026worldscapemoe} & 57.55 & 41.18 & 59.77 & 73.04 & 23.21 & 5.80 &
63.52 & 87.05 & 7.74 & 71.43 & 61.96 & 33.25 & 93.39 & 83.60 & 69.20 & 89.10 \\
PF-Cosmos \citep{zhang2026physisforcing} & 51.46 & 45.84 & 58.35 & 40.49 & 21.29 & 5.74 & 61.68 & 79.79 & 9.53 & 63.33 & 58.20 & 9.32 & 89.56 & 86.80 & 53.80 & 88.18 \\
Wan2.2 \citep{wanteam2025wan} & 49.21 & 42.87 & 46.18 & 59.61 & 17.86 & 5.06 & 57.75 & 73.39 & 19.11 & 62.04 & 44.32 & 14.15 & 86.73 & 75.88 & 45.52 & 87.61 \\
GigaWorld-0 \citep{gigaworldteam2025} & 48.06 & 43.30 & 50.57 & 44.41 & 22.59 & 5.79 & 62.73 & 79.77 & 3.60 & 56.11 & 47.46 & 15.24 & 77.22 & 74.98 & 51.22 & 85.92 \\
Vidar \citep{feng2025vidar} & 47.13 & 42.75 & 45.33 & 54.76 & 18.16 & 5.60 & 63.63 & 69.95 & 10.96 & 53.36 & 43.04 & 16.99 & 81.55 & 67.70 & 45.96 & 87.24 \\
IRASim \citep{zhu2024irasim} & 44.97 & 31.43 & 45.18 & 62.15 & 18.44 & 5.49 & 59.64 & 65.70 & 10.44 & 48.02 & 37.88 & 22.68 & 89.68 & 51.56 & 38.18 & 88.03 \\
\midrule
\rowcolor{red!5} KineWorld (Ours)$^\dagger$ & 68.95 & 44.47 & 59.80 & 97.00 & 23.93 & 11.99 & 64.15 & 91.90 & 27.38 & 87.74 & 82.40 & 70.57 & 99.65 & 97.40 & 86.60 & 89.28 \\
\bottomrule
\end{tabular}%
}
\label{tab:full_metrics}
\end{table*}
\begin{table*}[t]
\appendixtablestyle
\caption{\textbf{Complete Single-View Local Diagnostics.} Additional components for Table~\appmainref{tab:robotwin_singleview_diagnostics}; DYN and FL retain the local evaluator scale.}
\setlength{\tabcolsep}{0.5mm}
\begin{appendixtabular}{l YYYYYYYYYY}
\toprule
$\operatorname{Method}$ & $\operatorname{AES}\uparrow$ & $\operatorname{IMG}\uparrow$ & $\operatorname{BG}\uparrow$ & $\operatorname{SC}\uparrow$ & $\operatorname{PC}\uparrow$ & $\operatorname{INT}\uparrow$ & $\operatorname{PER}\uparrow$ & $\operatorname{IF}\uparrow$ & $\operatorname{DYN}$ & $\operatorname{FL}$ \\
\midrule
Wan2.2-TI2V-5B \citep{wanteam2025wan} & 42.10 & 44.42 & 85.17 & \textbf{80.90} & 0.96 & 68.68 & 88.52 & 76.00 & -- & -- \\
\rowcolor{red!5} KineWorld & \textbf{45.26} & \textbf{61.21} & \textbf{87.19} & 75.53 & \textbf{16.43} & \textbf{79.20} & \textbf{97.30} & \textbf{85.84} & 0.53 & 1.88 \\
\bottomrule
\end{appendixtabular}
\label{tab:local_singleview_full1000}
\end{table*}

\begin{figure}[!htbp]
    \centering
    \begin{minipage}[c]{0.55\linewidth}
        \centering
        \includegraphics[width=\linewidth]{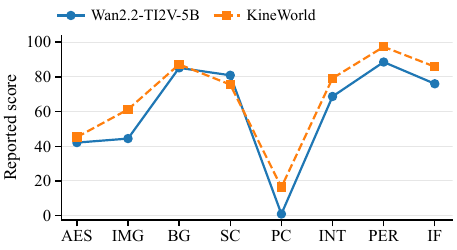}
    \end{minipage}\hfill
    \begin{minipage}[c]{0.42\linewidth}
        \small
        \caption{\small\raggedright\textbf{Single-view local diagnostic profiles.} Reported AES--IF values from Table~\ref{tab:local_singleview_full1000}, retaining the normalized score scale and metric abbreviations in Table~\ref{tab:full_metrics}. Scores are compared within each metric. The profiles describe the reported local systems rather than a checkpoint-matched KTL or TAWD ablation.}
        \label{fig:singleview_local_diagnostics}
    \end{minipage}
\end{figure}

\paragraph{Local Evaluation.}
Table~\ref{tab:local_singleview_full1000} supplements Table~\appmainref{tab:robotwin_singleview_diagnostics} with the reported 1,000-video local diagnostic profile. AES--IF use the displayed normalized score scale, while DYN and FL use the raw local evaluator scale. A dash denotes an unavailable measurement. These local components are analyzed separately from the normalized reference profile in Table~\ref{tab:full_metrics}.
Figure~\ref{fig:singleview_local_diagnostics} visualizes the eight shared components: KineWorld has higher reported scores on seven, with lower Subject Consistency. This descriptive profile does not isolate the effect of changing the training objective.

\subsubsection{Metric Definitions and Score Aggregation}
\label{app:metric_aggregation}

\paragraph{EWMScore-P and Component Scores.}
We use the 15-component perceptual score EWMScore-P, abbreviated as EWM in
the tables, following the Track~1 scoring convention
\citep{shang2026worldarena2}.\footnote{The
\href{https://huggingface.co/spaces/WorldArena/WorldArena2.0/blob/7f1bc898ab88a950da1aabb9d08f235bd2e743cc/app.py}{official display code}
names the score EWMScore-P; the
\href{https://huggingface.co/spaces/WorldArena/WorldArena2.0/blob/7f1bc898ab88a950da1aabb9d08f235bd2e743cc/src/data_loader.py}{pinned aggregation implementation}
defines its 15 components and six groups.}
This is a benchmark metric, not a new KineWorld score or a RoboTwin~2.0
task success rate. Let $m_k$ be component $k$ after its
metric-specific normalization, applicable protocol corrections, and
dataset-level aggregation. The displayed component score is
\begin{equation}
    S_k=Q_2(100m_k),
    \qquad k=1,\ldots,15,
    \label{eq:metric_score_scale}
\end{equation}
where $Q_2$ rounds to two decimal places, as in the pinned public display
implementation. The factor 100 changes the reporting scale; it does not
turn a similarity or quality score into a success percentage.

\paragraph{Metric-Specific Evaluation and Normalization.}
The components measure image quality (IMG), aesthetic quality (AES),
video-representation similarity (JEPA), dynamic degree (DYN), optical-flow
activity (FL), motion smoothness (MS), subject consistency (SC), background
consistency (BG), photometric consistency (PC), interaction quality (INT),
trajectory accuracy (TA), depth accuracy (DEP), perspectivity (PER),
instruction following (IF), and semantic alignment (SA). Their feature
extractors, motion estimators, and semantic judges are those of the
benchmark, rather than KineWorld-specific evaluators.

The public implementation uses metric-specific score mappings rather than
a single normalization fitted across the compared models.\footnote{See
the pinned
\href{https://github.com/WorldArena2/WorldArena-2.0/tree/222f7b9bd225639e88e37f605a1067e9d2621801/video_quality_ood}{evaluation implementation}.
Here, $r$ denotes a component evaluator's returned value, not necessarily
an untransformed pixel or trajectory error.}
For example, IMG and AES divide their native quality scores by 100 and
10, respectively; INT, PER, and IF divide the corresponding judge scores
by 5. Writing $C(x)=\min(1,\max(0,x))$, the released empirical
normalization functions include
\begin{equation}
\begin{aligned}
    \mathcal N_{\mathrm{PC}}(r)&=C\!\left(\frac{r-0.1257}{6.7899-0.1257}\right),
    &\mathcal N_{\mathrm{MS}}(r)&=C\!\left(\frac{r}{2.6413}\right),\\
    \mathcal N_{\mathrm{FL}}(r)&=C\!\left(\frac{r-0.0531}{8.9414-0.0531}\right),
    &\mathcal N_{\mathrm{TA}}(r)&=C\!\left(\frac{r}{40.8540}\right),\\
    \mathcal N_{\mathrm{DEP}}(r)&=
        1-C\!\left(\frac{r-0.2228}{4.3711-0.2228}\right).
\end{aligned}
\label{eq:metric_normalization}
\end{equation}
The PC implementation uses the internal key
\texttt{photometric\_smoothness}. These mappings are applied to raw
component outputs, not reapplied to already normalized table entries.
JEPA is a collection-level score: the released base evaluator converts its
JEDi distance $d_{\mathrm{JEDi}}$ through
$\mathcal N_{\mathrm{JEPA}}(d_{\mathrm{JEDi}})
=\exp(-0.4d_{\mathrm{JEDi}})$.
Thus, the 15 components must not all be interpreted as independent
per-video scores followed by the same averaging operation.

\paragraph{Underlying Component Evaluators.}
The following definitions distinguish the raw evaluator $r_k$ from its
normalization $\mathcal N_k$ above. Let $I_{1:T}$ and $I^{\mathrm{GT}}_{1:T}$
denote generated and reference frames on each evaluator's sampling grid.
All feature networks, prompts, image preprocessing, and trajectory
extractors are fixed benchmark components, not trained with KineWorld.
For learned judges, a scoring-function definition specifies the actual
inputs and reduction; it is not a closed-form replacement for the network.

\paragraph{Image, Aesthetic, and Semantic Quality.}
Image quality uses MUSIQ--SPAQ, while aesthetic quality uses a fixed
linear head on unit-normalized CLIP ViT-L/14 image features:
\begin{equation}
\begin{aligned}
 r_{\mathrm{IMG}}&=\frac1T\sum_t f_{\mathrm{MUSIQ}}(I_t),
 &r_{\mathrm{AES}}&=\frac1T\sum_t (w^\top z_t+b),\\
 z_t&=\frac{f_{\mathrm{CLIP\mbox{-}L/14}}(I_t)}
 {\|f_{\mathrm{CLIP\mbox{-}L/14}}(I_t)\|_2}.
\end{aligned}
\label{eq:learned_image_metrics}
\end{equation}
Their normalized values are $r_{\mathrm{IMG}}/100$ and
$r_{\mathrm{AES}}/10$. Semantic alignment compares the generated and
reference videos' \texttt{General} captions, not a video embedding with
the commanded instruction:
\begin{equation}
 r_{\mathrm{SA}}=\cos\!\left(f^{\mathrm{text}}_{\mathrm{CLIP}}(c(I)),
                         f^{\mathrm{text}}_{\mathrm{CLIP}}(c(I^{\mathrm{GT}}))\right).
\label{eq:caption_semantic_alignment}
\end{equation}
Here $c$ is the benchmark's Qwen-generated caption and the CLIP text
tokenizer truncates at 77 tokens. No additional cosine remapping is
introduced. For $k\in\{\mathrm{INT},\mathrm{PER},\mathrm{IF}\}$, the
normalized score is $\mathcal J_k(\mathcal X_k)/5$, where
$\mathcal J_k$ is the corresponding output of the shared Qwen3-VL
judge and $\mathcal X_k$ contains uniformly sampled RGB frames and
the task instruction.
These learned ratings measure interaction, perspective, and instruction
following; they are not binary success indicators.

\paragraph{Motion Activity and Temporal Consistency.}
For adjacent-frame RAFT flow $F_t$, write $a_t$ for the mean of the top
5\% pixel magnitudes and $\tau=6\min(H,W)/256$. The dynamic-degree and
raw flow statistics are
\begin{equation}
\begin{aligned}
 r_{\mathrm{DYN}}&=\frac1{T-1}\sum_t
   \sigma\!\left(5\left(\frac{a_t}{\tau}-1\right)\right),\\
 r_{\mathrm{FL}}&=\frac1{(T-1)|\Omega|}
   \sum_{t,p\in\Omega}\|F_t(p)\|_2.
\end{aligned}
\label{eq:flow_activity_metrics}
\end{equation}
$\sigma$ is the logistic function and $\Omega$ includes the evaluator's
RAFT padding. Dynamic degree subsamples MP4s near 8 fps, whereas flow
score uses adjacent frames; $T$ therefore refers to the selected grid
for the particular evaluator. Neither statistic alone measures whether
the motion follows the command.

Let $[x]_+=\max(x,0)$ and
$g(d)=d$ for $d\leq0.1213$, with $g(d)=1$ otherwise. The single-view
subject and background scores use
\begin{equation}
\begin{aligned}
 C_f(I)&=\frac1{2(T-1)}\sum_{t=2}^T
 \left([\cos(f(I_{t-1}),f(I_t))]_+
       +[\cos(f(I_1),f(I_t))]_+\right),\\
 r_{\mathrm{SC}}&=g(r_{\mathrm{DYN}})C_{\mathrm{DINO\mbox{-}B/16}}(I),
 \qquad
 r_{\mathrm{BG}}=g(r_{\mathrm{DYN}})C_{\mathrm{CLIP\mbox{-}B/32}}(I).
\end{aligned}
\label{eq:temporal_feature_consistency}
\end{equation}
These are feature-consistency scores with the released low-motion
penalty, not region-masked pixel errors.

Motion smoothness reconstructs a midpoint with VFIMamba. Using
zero-based indices, define $a_j=\operatorname{MAE}(I_{2j},I_{2j+2})$
on the 0--255 RGB scale and $\mathcal J=\{j:a_j\geq1\}$:
\begin{equation}
 r_{\mathrm{MS}}=\frac1{|\mathcal J|}\sum_{j\in\mathcal J}
 \operatorname{SSIM}\!\left(\operatorname{VFIMamba}(I_{2j},I_{2j+2}),I_{2j+1}\right)
 \log(1+a_j).
\label{eq:midpoint_motion_smoothness}
\end{equation}
The implementation quantizes the predicted midpoint before SSIM and
returns zero when $\mathcal J$ is empty. Photometric consistency
(the implementation key is \texttt{photometric\_smoothness}) instead
uses SEA-RAFT forward/backward cycle error. For cropped flow fields
$F_t,B_t$ and rounded, clipped lookup $\mathcal W(p,F_t)$,
\begin{equation}
\begin{aligned}
 e_t&=\frac1{|\Omega_c|}\sum_{p\in\Omega_c}
      \|F_t(p)+B_t(\mathcal W(p,F_t))\|_2,\\
 r_{\mathrm{PC}}&=\frac{g(d_{\mathrm{SEA}})}{(T-1)^{-1}\sum_t e_t}.
\end{aligned}
\label{eq:cycle_photometric_score}
\end{equation}
$\Omega_c$ is the central $(H-30)\times(W-30)$ crop, and
$d_{\mathrm{SEA}}$ applies the dynamic-degree construction to SEA-RAFT
flows. The released score is reciprocal cycle error, not RGB warping
error. No additional stabilization is introduced in the reciprocal.

\paragraph{Trajectory and Depth Accuracy.}
The trajectory evaluator selects the valid reference track $k^\star$
with the greatest spatial diameter and aligns it with the corresponding
generated track by FastDTW. For alignment path $\pi$,
\begin{equation}
 r_{\mathrm{TA}}=\left[
 \frac1{|\pi|}\sum_{(u,v)\in\pi}
 \|P^{\mathrm{gen}}_{k^\star,u}-P^{\mathrm{GT}}_{k^\star,v}\|_2
 \right]^{-1}.
\label{eq:trajectory_raw_score}
\end{equation}
The released tracker interpolates missing points and rejects tracks
with more than 90\% missing samples; an invalid selected generated
track receives zero. This is an image-trajectory score, not joint-space
control error. Depth accuracy estimates depths from both generated and
reference RGB and computes median-scale-aligned relative error:
\begin{equation}
\begin{aligned}
 a_D&=\frac{\operatorname{med}(D^{\mathrm{GT}})}
                 {\operatorname{med}(D^{\mathrm{gen}})},\\
 r_{\mathrm{DEP}}&=\frac1{|\mathcal V|}\sum_{(t,p)\in\mathcal V}
 \frac{|a_DD^{\mathrm{gen}}_t(p)-D^{\mathrm{GT}}_t(p)|}
      {D^{\mathrm{GT}}_t(p)+10^{-6}}.
\end{aligned}
\label{eq:depth_raw_error}
\end{equation}
Here $\mathcal V=\{(t,p):D^{\mathrm{GT}}_t(p)>10^{-3}\}$;
the scale uses video-wide medians, defaults to one for a nonpositive
generated median, and an empty mask falls back to all pixels. The
lower-is-better error is inverted by $\mathcal N_{\mathrm{DEP}}$.
Depth estimates, preprocessing, and sampling remain those of the
released evaluator rather than simulator ground-truth depth.

\paragraph{Distributional JEPA Similarity.}
For generated and reference V-JEPA features $x_i,y_j\in\mathbb R^d$,
the bundled JEDi evaluator uses the biased squared polynomial-kernel MMD:
\begin{equation}
\begin{aligned}
 k(x,y)&=(x^\top y/d)^2,\\
 D_{\mathrm{JEDi}}&=100\left[
 \frac1{n^2}\sum_{i,i'}k(x_i,x_{i'})
 +\frac1{m^2}\sum_{j,j'}k(y_j,y_{j'})
 -\frac2{nm}\sum_{i,j}k(x_i,y_j)\right],\\
 m_{\mathrm{JEPA}}&=\exp(-0.4D_{\mathrm{JEDi}}).
\end{aligned}
\label{eq:jedi_distribution_score}
\end{equation}
Diagonal kernel terms are included and no square root is applied.
This is a collection-level distribution comparison, not an average of
per-episode embedding distances. The raw definitions above describe
per-video component records, except for the collection-level JEPA
score, rather than auxiliary aggregate return values. Source-specific
preprocessing, rounding, and failure behavior are part of the evaluator,
including its depth-input byte conversion; the equations do not imply
changes to that implementation.

\paragraph{Reference-Capped Motion Metrics.}
Under the revised motion protocol, let $a_{nk}$ and $a^{\mathrm{GT}}_{nk}$
be the normalized scores for generated video $n$ and its matched
ground-truth video. For $k\in\{\mathrm{DYN},\mathrm{FL},\mathrm{MS}\}$,
the corrected per-video score is
\begin{equation}
    \widetilde a_{nk}=\min\!\left(a_{nk},a^{\mathrm{GT}}_{nk}\right).
    \label{eq:reference_capped_motion}
\end{equation}
This correction is applied before the benchmark's component aggregation,
not by capping a dataset mean after aggregation. It prevents motion beyond
the reference level from automatically receiving additional credit.
The other components are unchanged by this motion-specific
revision.\footnote{The
\href{https://github.com/WorldArena2/WorldArena-2.0/blob/222f7b9bd225639e88e37f605a1067e9d2621801/video_quality_ood/WorldArena/__init__.py}{official evaluator}
specifies the per-video matched-reference cap.}
Public profiles retain the component scores supplied by their source
snapshots, without retrospective recalibration. No additional difficulty
or out-of-distribution multiplier is applied when aggregating these
components in the paper.

\paragraph{Six-Dimension Aggregation.}
Using the abbreviations above, the six dimension scores are group means:
\begin{equation}
\begin{aligned}
    G_{\mathrm{Vis}}&=\frac{S_{\mathrm{IMG}}+S_{\mathrm{AES}}+S_{\mathrm{JEPA}}}{3},\\
    G_{\mathrm{Mot}}&=\frac{S_{\mathrm{DYN}}+S_{\mathrm{FL}}+S_{\mathrm{MS}}}{3},\\
    G_{\mathrm{Cont}}&=\frac{S_{\mathrm{SC}}+S_{\mathrm{BG}}+S_{\mathrm{PC}}}{3},\\
    G_{\mathrm{Phys}}&=\frac{S_{\mathrm{INT}}+S_{\mathrm{TA}}}{2},\\
    G_{\mathrm{3D}}&=\frac{S_{\mathrm{DEP}}+S_{\mathrm{PER}}}{2},\\
    G_{\mathrm{Ctrl}}&=\frac{S_{\mathrm{IF}}+S_{\mathrm{SA}}}{2}.
\end{aligned}
\label{eq:ewm_dimension_scores}
\end{equation}
These correspond to Visual Quality, Motion Quality, Content Consistency,
Physics Adherence, 3D Accuracy, and Controllability. The displayed
dimension values are $Q_2(G_g)$.

\paragraph{Overall Perceptual Score.}
For a complete 15-component profile, the overall score is
\begin{equation}
\begin{aligned}
    \mathrm{EWMScore\mbox{-}P}
    &=Q_2\!\left(\frac{1}{15}\sum_{k=1}^{15}S_k\right)\\
    &=Q_2\!\left(
    \frac{3G_{\mathrm{Vis}}+3G_{\mathrm{Mot}}+3G_{\mathrm{Cont}}
    +2G_{\mathrm{Phys}}+2G_{\mathrm{3D}}+2G_{\mathrm{Ctrl}}}{15}
    \right).
\end{aligned}
\label{eq:ewmscore_p}
\end{equation}
Every component has weight $1/15$. The six dimensions are therefore
\emph{not} equally weighted: their group sizes are $3,3,3,2,2,2$.
The second expression uses the group means before their display rounding.
For illustration, the 15 displayed KineWorld components in
Table~\ref{tab:full_metrics} sum to 1034.26, giving
$Q_2(1034.26/15)=68.95$.

\paragraph{Local Diagnostic Scales.}
Raw local DYN and FL measurements in
Table~\ref{tab:local_singleview_full1000} are not substituted for the
normalized components in Equation~\ref{eq:ewmscore_p}. A partial component
profile is not reported as a complete 15-metric EWMScore-P.
The metric name specifies the scoring convention, not whether a result
is an official submission; the source and comparison scope remain those
in Appendix~\ref{app:result_provenance}.

\subsubsection{TriWorldBench Metrics and TWB-Score}
\label{app:twb_metrics}

\paragraph{Protocol and Reporting Scale.}
We use the official name TWB-Score for the multi-view aggregate
\citep{triworldbench2026}. It is distinct from the single-view
EWMScore-P and is not a task success rate. TriWorldBench evaluates
synchronized head, left-wrist, and right-wrist videos using 19 component
signals. Component names and reductions follow the pinned
implementation, including its task- and consistency-dependent visual
quality penalty.\footnote{See the
\href{https://github.com/TriWorldBench/TriWorldBench/blob/0d3603ec48df2448bd29d716c8de8cb6b45d3114/scripts/lib/eval_common.py}{TriWorldBench aggregation code},
\href{https://github.com/TriWorldBench/TriWorldBench/blob/0d3603ec48df2448bd29d716c8de8cb6b45d3114/metrics/_common/protocol_0710.py}{score transformations},
and
\href{https://github.com/TriWorldBench/TriWorldBench/blob/0d3603ec48df2448bd29d716c8de8cb6b45d3114/config/hyperparameters.yaml}{protocol configuration}.}
Let $b_{nk}$ denote the processed score of component $k$ for
episode $n$, after its prescribed camera reduction. Under the default
equal-episode reduction, the component summary is
\begin{equation}
    S^{\mathrm{mv}}_k=100\,\frac{1}{|\mathcal E_k|}
    \sum_{n\in\mathcal E_k}b_{nk},
    \qquad k=1,\ldots,19,
    \label{eq:twb_component_reduction}
\end{equation}
where $\mathcal E_k$ is the set of episodes with a valid component
output. A zero score is not a missing value. The implementation also
supports an optional frame-bucket weighting mode; it is not implicit
in the equal-episode expression above. Presentation rounding follows
the specified reduction; intermediate rating quantization and numerical
rounding remain part of the component evaluators.

\paragraph{Pixel Fidelity and Cross-View Judgments.}
For paired frames $X,Y\in[0,255]^{H\times W\times3}$, the benchmark
first evaluates frame-level PSNR and SSIM:
\begin{equation}
\begin{aligned}
 \operatorname{MSE}(X,Y)&=\frac{\|X-Y\|_F^2}{3HW},
 &\operatorname{PSNR}(X,Y)&=10\log_{10}\frac{255^2}{\operatorname{MSE}(X,Y)},\\
 \operatorname{SSIM}(X,Y)&=
 \frac{(2\mu_X\mu_Y+C_1)(2\sigma_{XY}+C_2)}
 { (\mu_X^2+\mu_Y^2+C_1)(\sigma_X^2+\sigma_Y^2+C_2)}.
\end{aligned}
\label{eq:twb_pixel_metrics}
\end{equation}
SSIM uses an $11\times11$ Gaussian window with standard deviation 1.5,
$C_1=(0.01\cdot255)^2$, and $C_2=(0.03\cdot255)^2$, followed by
spatial, channel, and frame averaging. PSNR is averaged in dB over the
evaluator's finite paired-frame values; exact matches have infinite
PSNR, retained when all selected values are infinite. Let
$\langle\cdot\rangle_v$ be the equal mean over the three cameras. The
reported normalized PSNR is
\begin{equation}
 b_{n,\mathrm{NPSNR}}=
 \left\langle C\!\left(1-\exp(-\operatorname{PSNR}_{nv}/20)\right)\right\rangle_v,
 \qquad
 b_{n,\mathrm{SSIM}}=\langle\operatorname{SSIM}_{nv}\rangle_v.
\label{eq:twb_normalized_psnr}
\end{equation}
NPSNR is therefore not PSNR in dB. Both metrics compare each predicted
camera with its own reference, rather than directly comparing pixels
from different camera poses.

The dedicated VLM consistency evaluators use STATE-selected phase
samples, excluding idle and approach phases. For valid sample sets
$\mathcal P_{n,j}$, their released score reductions are
\begin{equation}
\begin{aligned}
 b_{n,01}&=\frac1{|\mathcal P_{n,01}|}\sum_{p\in\mathcal P_{n,01}}
               \frac{s^{01}_p+o^{01}_p}{10},\\
 b_{n,02}&=\frac1{|\mathcal P_{n,02}|}\sum_{p\in\mathcal P_{n,02}}\frac{r^{02}_p}{5},\\
 b_{n,03}&=\frac1{|\mathcal P_{n,03}|}\sum_{p\in\mathcal P_{n,03}}
               \frac{s^{03}_p+o^{03}_p}{10}.
\end{aligned}
\label{eq:twb_vlm_consistency}
\end{equation}
Here $s_p,o_p$ are the parsed state/object ratings and $r_p$ is the
parsed overall rating produced by the fixed judge prompts. The identifiers 01--03
refer to the executable evaluator variants; the result tables retain
the source's VLM-I/II/III labels. Variant 01 uses a calibrated
three-camera strip, and variants 02--03 use the head and active-wrist
views. Variant 01 supports legacy 0--100 responses; variant 02
quantizes ratings to half-points, and variant 03 to integers on the
1--5 scale. Samples are averaged without phase-duration
weighting. For VQA, let $a_n$ be the fraction of correctly answered
reference questions and $u_n$ the view-check score:
\begin{equation}
 b_{n,\mathrm{VQA}}=
 \begin{cases}a_n,&u_n=1,\\a_n/2,&u_n\ne1.\end{cases}
\label{eq:twb_vqa}
\end{equation}
These are fixed-model judgments of supplied evidence, not analytic
geometric-consistency guarantees.

\paragraph{Task, Geometry, and Motion Components.}
IF uses the head-view judge rating divided by five. SA uses the
head-view caption cosine in Equation~\ref{eq:caption_semantic_alignment}.
JEPA uses Equation~\ref{eq:jedi_distribution_score} on the evaluation
collection; its single global value is repeated in the episode rows,
so it is not an independently varying episode score.
For INT and PER, normalized camera ratings $a_v=\mathcal J_v/5$ are
reduced by a robust mean. Let $m=\operatorname{median}_v a_v$ and
$d=\operatorname{median}_v|a_v-m|$. Retain cameras satisfying
$|a_v-m|\leq\max(0.25,2.5d)$, adding the closest cameras if needed to
retain at least two, and average the retained ratings.

FL applies $\mathcal N_{\mathrm{FL}}$ to each camera's mean flow magnitude
before equal-camera averaging. TA uses the head-view reciprocal
FastDTW distance in Equation~\ref{eq:trajectory_raw_score}, followed
by $C(r_{\mathrm{TA}}/40.854)$; an invalid predicted track receives
zero and an exact match saturates at one. STATE is distinct from the
single-view DYN score. Let $m_{nv}$ be the fraction of frames marked
moving for camera $v$, and let $d_{nv}$ be its RAFT-based dynamic
degree from Equation~\ref{eq:flow_activity_metrics}. With $\theta=0.1213$,
\begin{equation}
\begin{aligned}
 u(d)&=\begin{cases}e^{d-\theta},&d<\theta,\\1,&d\geq\theta,\end{cases}
 &z(d)&=\begin{cases}e^{-d},&d>\theta,\\1,&d\leq\theta,\end{cases}\\
 b_{n,\mathrm{STATE}}&=
 \left\langle m_{nv}C(d_{nv})u(d_{nv})+(1-m_{nv})z(d_{nv})\right\rangle_v.
\end{aligned}
\label{eq:twb_state_alignment}
\end{equation}
Wrist motion labels follow the corresponding arm; the head label uses
the logical OR of the two arms. The released moving-phase set includes
approach, grasp closing, manipulation, handover motion, action execution,
adjustment, and release opening. STATE mixes expected moving and static
fractions; it is not per-frame action classification accuracy.

\paragraph{Temporal Components and Visual-Quality Penalty.}
The temporal components use the feature and flow-cycle statistics
defined above, with the following multi-view reductions:
\begin{equation}
\begin{aligned}
 b_{n,\mathrm{SC}}&=\langle C_{\mathrm{DINO}}(I_{nv})\rangle_v,\\
 b_{n,\mathrm{BG}}&=\langle g(d_{nv})C_{\mathrm{CLIP}}(I_{nv})\rangle_v,\\
 b_{n,\mathrm{PS}}&=\left\langle
 \mathcal N_{\mathrm{PC}}\!\left(g(d^{\mathrm{SEA}}_{nv})/E^{\mathrm{cycle}}_{nv}\right)
 \right\rangle_v.
\end{aligned}
\label{eq:twb_temporal_components}
\end{equation}
$E^{\mathrm{cycle}}_{nv}$ is the mean cropped forward/backward cycle
error from Equation~\ref{eq:cycle_photometric_score}. The pinned
multi-view executor disables the legacy dynamic-degree multiplier for
SC, but retains it for BG and the flow-cycle score. Thus the single-view
SC formula cannot simply be reused without this distinction.

Image and aesthetic scores first average the normalized learned
frame scores over each view and then over cameras. For
$Q\in\{\mathrm{IMG},\mathrm{AES}\}$, the score entering the final CSV is
\begin{equation}
\begin{aligned}
 c_n&=\frac1{|\mathcal A_n|}\sum_{j\in\mathcal A_n}b_{n,j},
 \qquad \mathcal A_n\subseteq\{01,02,03\},\\
 b_{n,Q}&=C(b^{\mathrm{raw}}_{n,Q})
 \exp\!\left[\tfrac12\big(C(b_{n,\mathrm{TA}})-1\big)
             +\tfrac12\big(C(c_n)-1\big)\right].
\end{aligned}
\label{eq:twb_visual_penalty}
\end{equation}
Here $\mathcal A_n$ contains valid dedicated consistency outputs.
Penalties are applied before episode averaging; an empty $\mathcal A_n$
or missing raw quality or TA excludes that quality value rather than
assigning zero. Thus, multi-view IMG/AES are not interchangeable with
unpenalized single-view scores.\footnote{Native component
implementations and fixed-model routing are provided in
\href{https://github.com/TriWorldBench/TriWorldBench/tree/0d3603ec48df2448bd29d716c8de8cb6b45d3114/metrics}{the pinned metric modules};
the equations summarize those implementations without changing them.}

\paragraph{Six Multi-View Dimensions.}
Using the components in Table~\ref{tab:robotwin_multiview_components},
the dimension scores are
\begin{equation}
\begin{aligned}
    G^{\mathrm{mv}}_{\mathrm{Tri}}
    &=\frac{S^{\mathrm{mv}}_{\mathrm{NPSNR}}+S^{\mathrm{mv}}_{\mathrm{SSIM}}
      +S^{\mathrm{mv}}_{\mathrm{VLM\mbox{-}I}}+S^{\mathrm{mv}}_{\mathrm{VLM\mbox{-}II}}
      +S^{\mathrm{mv}}_{\mathrm{VLM\mbox{-}III}}+S^{\mathrm{mv}}_{\mathrm{VQA}}}{6},\\
    G^{\mathrm{mv}}_{\mathrm{Task}}
    &=\frac{S^{\mathrm{mv}}_{\mathrm{IF}}+S^{\mathrm{mv}}_{\mathrm{SA}}
      +S^{\mathrm{mv}}_{\mathrm{JEPA}}}{3},\\
    G^{\mathrm{mv}}_{\mathrm{Phys}}
    &=\frac{S^{\mathrm{mv}}_{\mathrm{INT}}+S^{\mathrm{mv}}_{\mathrm{PER}}}{2},\\
    G^{\mathrm{mv}}_{\mathrm{Mot}}
    &=\frac{S^{\mathrm{mv}}_{\mathrm{STATE}}+S^{\mathrm{mv}}_{\mathrm{FL}}
      +S^{\mathrm{mv}}_{\mathrm{TA}}}{3},\\
    G^{\mathrm{mv}}_{\mathrm{Temp}}
    &=\frac{S^{\mathrm{mv}}_{\mathrm{SC}}+S^{\mathrm{mv}}_{\mathrm{BG}}
      +S^{\mathrm{mv}}_{\mathrm{PS}}}{3},\\
    G^{\mathrm{mv}}_{\mathrm{Vis}}
    &=\frac{S^{\mathrm{mv}}_{\mathrm{IMG}}+S^{\mathrm{mv}}_{\mathrm{AES}}}{2}.
\end{aligned}
\label{eq:multiview_highlighted_dimensions}
\end{equation}
These correspond to Tri-View Consistency, Task Alignment, Physical and
3D Coherence, Motion Quality, Temporal Consistency, and Visual Quality.
PS denotes Photometric Smoothness, not PSNR. The single- and multi-view
groups are different even when they share a short label such as Phys.
For KineWorld, the two highlighted multi-view dimensions give
$Q_2((37.55+88.92)/2)=63.24$ and
$Q_2((86.02+66.71+56.97)/3)=69.90$.

\paragraph{Overall TWB-Score.}
For a complete 19-component profile, the official component-mean rule is
\begin{equation}
\begin{aligned}
    \mathrm{TWB\mbox{-}Score}
    &=Q_2\!\left(\frac{1}{19}\sum_{k=1}^{19}S^{\mathrm{mv}}_k\right)\\
    &=Q_2\!\left(\frac{
      6G^{\mathrm{mv}}_{\mathrm{Tri}}+3G^{\mathrm{mv}}_{\mathrm{Task}}
      +2G^{\mathrm{mv}}_{\mathrm{Phys}}+3G^{\mathrm{mv}}_{\mathrm{Mot}}
      +3G^{\mathrm{mv}}_{\mathrm{Temp}}+2G^{\mathrm{mv}}_{\mathrm{Vis}}
      }{19}\right).
\end{aligned}
\label{eq:twb_score}
\end{equation}
The six dimensions are not equally weighted: their group sizes are
$6,3,2,3,3,2$. The single-view 15-component rule is not used here.
The displayed KineWorld components sum to 1041.61, giving
$Q_2(1041.61/19)=54.82$; computations from rounded component tables
can differ from those using the unrounded source outputs.
The implementation can summarize a selected subset of available
metrics, but such a partial summary is not a complete 19-component
profile. We retain the published aggregate values for external systems
rather than replacing them with recomputations from rounded entries.
Episode-paired uncertainty is treated separately in
Appendix~\ref{app:statistical_protocol}.

\subsection{More Details of KineWorld}
\label{app:objective_derivation}

\begin{figure}[t]
    \centering
    \includegraphics[width=\linewidth]{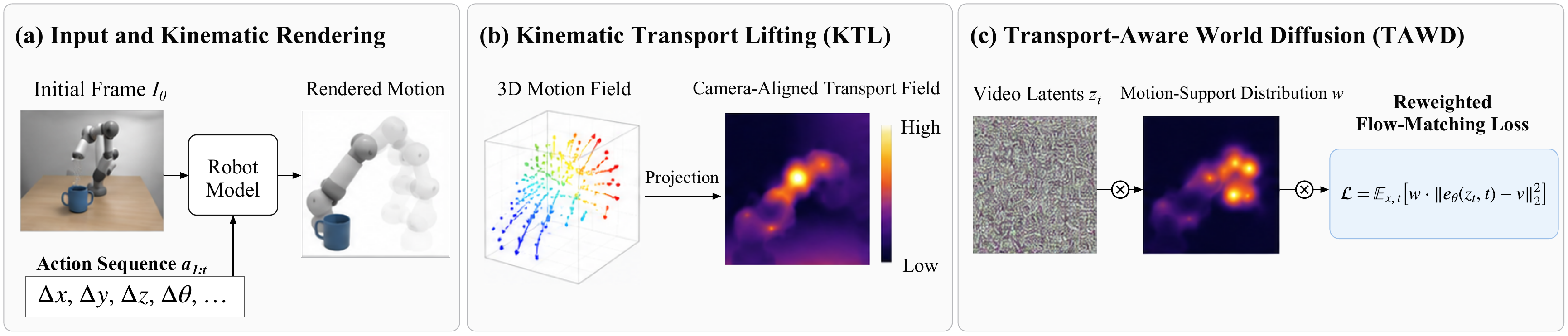}
    \caption{\textbf{Conceptual illustration of the KineWorld pipeline.} KTL renders robot states and uses RAFT to construct camera-aligned transport, which conditions the generator. TAWD detaches the transport path, aggregates calibrated activity on the latent grid, and converts it into a transport distribution $\pi_i$ and mean-one weights $w_i=N\pi_i$ for future-RGB flow-matching errors. TAWD changes training-time supervision only; it adds no inference steps or parameters.}
    \label{fig:appendix_pipeline_illustration}
\end{figure}

Figure~\ref{fig:appendix_pipeline_illustration} gives a conceptual view of the rendering, transport, and loss-weighting stages. We then provide the normalization properties and gradient boundary omitted from the main text. The derivation makes explicit that KineWorld reallocates a fixed supervision budget and recovers the uniform objective when no transport support is present.

Let $\mathcal P$ be the set of $N$ future RGB latent tokens and let $d_i\geq 0$ denote the calibrated transport support at token $i$. For $S_d=\sum_{j\in\mathcal P}d_j>\varepsilon_{\mathrm{sup}}$, KineWorld defines
\begin{equation}
\rho_i^{\mathrm{tr}}=\frac{d_i}{S_d},
\qquad
\pi_i=\frac{1-\alpha_\beta}{N}+\alpha_\beta\rho_i^{\mathrm{tr}},
\qquad
\alpha_\beta=\frac{\beta}{1+\beta}.
\label{eq:appendix_distribution}
\end{equation}
The mean-one weight is $w_i=N\pi_i$. Using the channel-averaged error from Equation~\appmainref{eq:token_error} and suppressing the outer expectation in Equation~\appmainref{eq:kine_loss} for brevity gives
\begin{equation}
\mathcal L_{\mathrm{Kine}}
=\frac{1}{N}\sum_{i\in\mathcal P}w_i e_i
=\sum_{i\in\mathcal P}\pi_i e_i,
\qquad
e_i=\frac{1}{C_z}\lVert v_{\theta,i}-v_i^*\rVert_2^2.
\label{eq:appendix_kine_loss}
\end{equation}

\paragraph{Fixed Aggregate Budget.}
Because $\sum_i\rho_i^{\mathrm{tr}}=1$, Equation~\ref{eq:appendix_distribution} gives
\begin{equation}
\sum_i\pi_i=1,
\qquad
\frac{1}{N}\sum_iw_i=1.
\label{eq:budget_conservation}
\end{equation}
Changing $\beta$ redistributes supervision while preserving the sum of token weights; it does not fix the realized loss or gradient norm. The uniform component remains non-zero for every finite $\beta$, so static appearance and contextual dynamics continue to receive supervision.

\paragraph{Zero-Motion Fallback.}
When $S_d\leq\varepsilon_{\mathrm{sup}}$, we set $\rho_i^{\mathrm{tr}}=1/N$, as in Equation~\appmainref{eq:transport_distribution}. Consequently, $\pi_i=1/N$ and $w_i=1$, exactly recovering uniform flow matching. This branch avoids numerical dependence on an arbitrary all-zero normalization.

\paragraph{Gradient Boundary.}
With disjoint generator and action-head parameters, and with action-head inputs restricted to the stopped features in Equation~\appmainref{eq:auxiliary_action_prediction}, the auxiliary loss satisfies
\begin{equation}
\nabla_{\theta_{\mathrm{wm}}}\mathcal L_{\mathrm{act}}=0.
\end{equation}
This equality requires that no shared parameter or additional differentiable path connects the auxiliary loss to the generator. Action-head parameters $\theta_{\mathrm{act}}$ remain trainable through this loss, but their gradient can be zero, for example at a stationary point.

\begin{figure}[t]
    \centering
    \includegraphics[width=\linewidth]{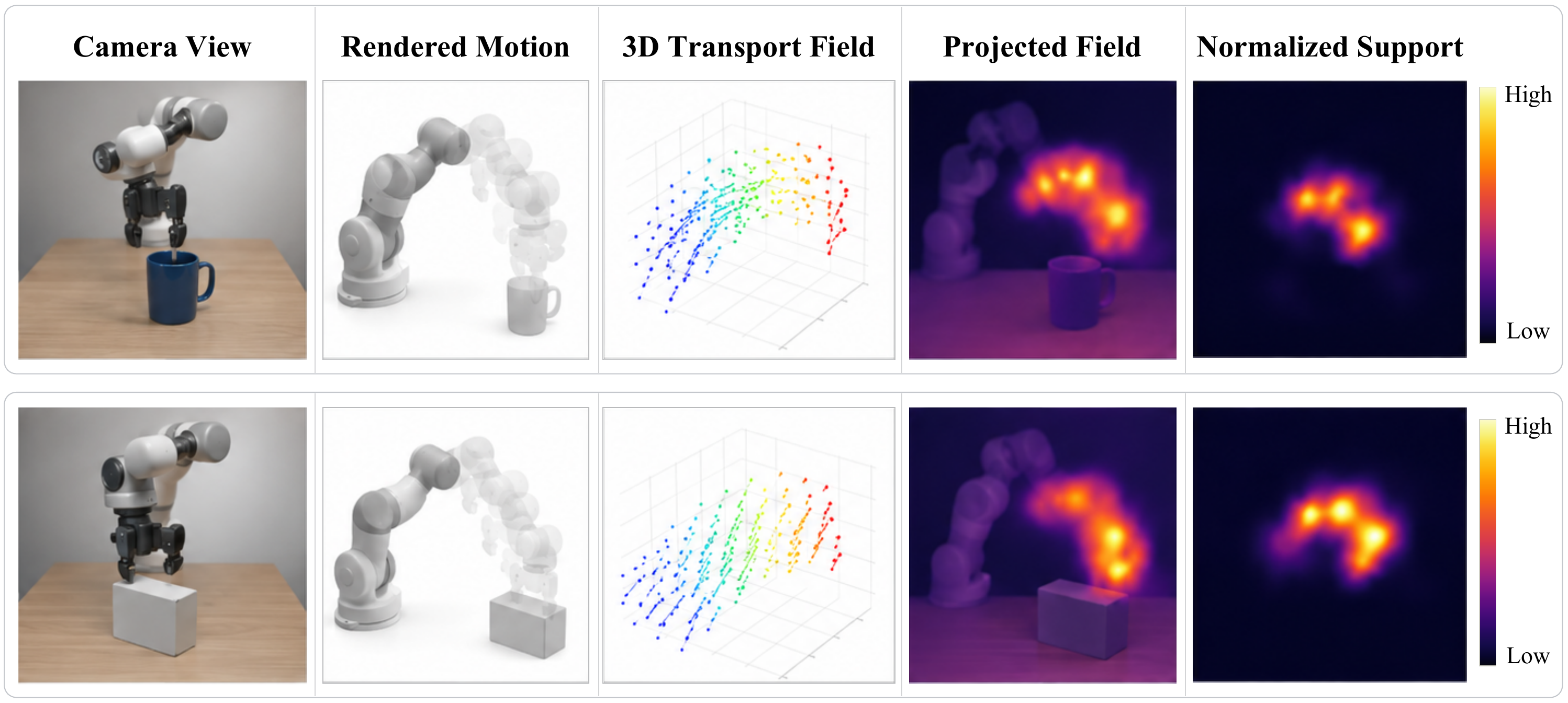}
    \caption{\textbf{Conceptual visualization of transport-support construction.} Two manipulation scenes illustrate camera views, rendered robot motion, schematic 3D fields, image-plane transport, and normalized support. KTL estimates transport using RAFT between rendered robot states rather than analytic 3D-field projection. The final column illustrates $\rho^{\mathrm{tr}}$, not the mean-one loss weights $w$. These panels are illustrative rather than measured experiment outputs.}
    \label{fig:transport_fields_illustration}
\end{figure}

\paragraph{Support Interpretation.}
The transport field is a renderer-derived, camera-aligned estimate of projected robot motion, not a sufficient statistic for the full action. Gripper semantics, contact forces, object state, and unobserved dynamics may not be recoverable from image-plane transport alone. KineWorld uses this estimate to decide where supervision is concentrated while language, the initial scene, and the generative model retain responsibility for the remaining dynamics. Figure~\ref{fig:transport_fields_illustration} conceptually illustrates the relation between projected robot motion and normalized transport support.

\begin{figure}[!htbp]
    \centering
    \begin{minipage}[t]{0.485\linewidth}
        \centering
        \includegraphics[width=\linewidth]{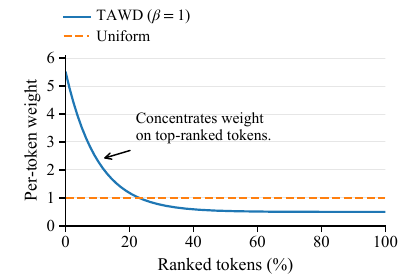}
        \par\smallskip{\small (a)}
    \end{minipage}\hfill
    \begin{minipage}[t]{0.485\linewidth}
        \centering
        \includegraphics[width=\linewidth]{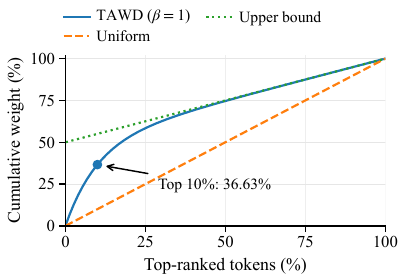}
        \par\smallskip{\small (b)}
    \end{minipage}
    \caption{\textbf{Analytical illustration of TAWD token-weight allocation.} For the constructed exponential support over $N=1000$ future RGB latent tokens at $\beta=1$, (a) shows descending mean-one weights and (b) their cumulative share relative to uniform weighting. The top 10\% receive 36.63\% of the total token weight. The green curve shows the general upper bound $(1+q)/2$ for $q>0$; the cumulative share has $C_0=0$. These curves illustrate the weighting rule, not measured training statistics or loss/gradient shares.}
    \label{fig:tawd_weight_allocation_theory}
\end{figure}

\paragraph{Illustrative Weight Allocation.}
Figure~\ref{fig:tawd_weight_allocation_theory} illustrates how TAWD redistributes a fixed token-weight budget. We construct $N=1000$ ordered future RGB latent tokens with $d_i=\exp[-10(i-1)/(N-1)]$ and apply Equation~\ref{eq:appendix_distribution} at $\beta=1$. For the top $k$ tokens in descending weight order, let $q=k/N$ and $C_k=\sum_{i=1}^{k}w_i/N$. Uniform weighting gives $C_k=q$, while TAWD satisfies $q\leq C_k\leq(1+q)/2$ for $k=1,\ldots,N$, with $C_0=0$. In this construction, the top 10\% receive 36.63\% of the total token weight, below the general upper bound of 55\%. This construction illustrates normalization and concentration, not an observed training distribution or a performance gain.

\subsection{Implementation Details}
\label{app:implementation}

\paragraph{Training Sequence.}
A reproducible KineWorld update consists of the following operations:
\begin{enumerate}
    \item integrate the control sequence from the initial embodiment configuration and sample $K=9$ visual states at stride $r=4$;
    \item render robot-only frames with the training camera and obtain source-grid forward flow between successive rendered states;
    \item mask flow by source-frame robot occupancy, clip displacement at $M_{\mathrm{px}}=25$, and construct the three-channel transport video;
    \item encode RGB and transport clips with the video-VAE interfaces, keeping the initial RGB and transport streams clean;
    \item downsample source-supported activity to the future-RGB latent grid and normalize it according to Equation~\ref{eq:appendix_distribution};
    \item predict the conditional flow field with the dual-stream denoiser and apply Equation~\ref{eq:appendix_kine_loss}; and
    \item update the optional action expert through detached generator features and add $\lambda_{\mathrm{act}}\mathcal L_{\mathrm{act}}$ to the scalar training objective.
\end{enumerate}

\paragraph{Matched-Comparison Protocol.}
Both runs start from the step-315 checkpoint and use 50 updates, a learning rate of $10^{-5}$ without warmup, and $\lambda_{\mathrm{act}}=1$. They share the optimizer reset, data order, seeds, KTL condition, sampler, chunking, interpolation, encoding, and evaluation settings. The only change is $\beta=0$ versus $\beta=1$ in the future-RGB loss.

\paragraph{Inference Path and Cost Accounting.}
TAWD modifies only the training objective, leaving the dual-stream generator and its 25 denoising steps unchanged. At inference, KTL converts each new action trajectory into a transport condition through robot rendering, RAFT estimation, and transport encoding. The reported FPS is measured with precomputed transport and excludes this preparation. Appendix~\ref{app:efficiency_audit} specifies the timing protocol.

\subsection{Audited Model Size and Runtime}
\label{app:efficiency_audit}

\paragraph{Parameter Accounting.}
We count instantiated denoiser tensor entries rather than checkpoint bytes and identify trainable parameters from the frozen continuation contract. The Wan2.2 DiT contains 4,999,787,712 parameters. The transport input/output stream adds 1,192,128 parameters, or 0.02384\% of the DiT, for an active single-view denoising path of 5,000,979,840 parameters (100.02384\% of the DiT). TAWD is a training objective and therefore adds zero learned parameters. The selected self-attention-LoRA continuation exposes 80,415,936 trainable parameters (1.60839\% of the DiT): 67,430,400 modulation parameters, 11,796,480 LoRA parameters, and 1,189,056 parameters in the checkpoint-version transport stream. The 3,072-parameter difference between this training contract and the instantiated inference stream is the inference revision's stream-identity embedding; rounding both footprints to two decimal places gives 1.19M transport and 80.42M trainable parameters. This trainable subset is not an additional component to be added to the denoiser total. An optional 778,125,326-parameter action expert (15.56317\% of the DiT) is stored in the joint training checkpoint, but the reported single-view loader ignores it; we consequently exclude it from the active denoiser size. The frozen video VAE and text encoder are also excluded so that the count follows the DiT-backbone convention.

\paragraph{Runtime Protocol.}
The reported timing profile covers 1,000 single-view episodes generated by 32 single-device workers across four nodes with 25 denoising steps. The outputs contain 230,908 frames (75--1,468 per episode), with mean and median episode times of 76.37~s and 61.60~s, respectively. Worker-normalized throughput is $\sum_n F_n/\sum_n t_n$, where $F_n$ is exported frame count and $t_n$ is episode wall time. Timing includes input loading, denoising, linear frame expansion, and $640\times480$ H.264 export, but excludes model initialization and offline transport construction. This measures exported-video throughput with precomputed transport. Videos are exported at 24~fps for playback.

\paragraph{Joint-Attention Cost.}
For $n_x$ RGB tokens and $n_u$ transport tokens with hidden width $d$, the quadratic term of joint self-attention scales as $(n_x+n_u)^2d$. Relative to RGB-only attention, this term increases by a factor of $(1+n_u/n_x)^2$, reaching $4\times$ for equal-length streams. This factor applies to the quadratic attention term rather than full-model runtime. Uniform CFM and TAWD share the same transport-conditioning architecture and therefore incur the same attention cost.

\subsection{Evaluation and Statistical Protocol}
\label{app:statistical_protocol}

The controlled-comparison protocol separates changes to future-RGB supervision from changes to initialization, training budget, or inference. The following estimator applies to episode-paired outputs under shared evaluation conditions; reference profiles are analyzed descriptively.

\paragraph{Primary Causal Contrast.}
Let $s_n^{\mathrm{KW}}$ and $s_n^{\mathrm{U}}$ be per-episode evaluator scores from KineWorld and the continued uniform control, generated with the same episode input and inference seed. Here $M=1000$ complete episode pairs. For metrics defined as episode-wise averages, the paired estimator is
\begin{equation}
\widehat{\Delta}=\frac{1}{M}\sum_{n=1}^{M}
\left(s_n^{\mathrm{KW}}-s_n^{\mathrm{U}}\right).
\end{equation}
We form 95\% percentile bootstrap intervals from 20,000 resamples of the $M$ complete episode pairs with replacement. All frames, cameras, and metrics from each episode remain grouped, and the same sampled indices are used for both methods. EWMScore-P is recomputed in each replicate using the prespecified component-aggregation rule before taking the paired difference. These intervals quantify episode-sampling uncertainty for the fixed checkpoint pair, rather than variability across independent training runs.

    \paragraph{Multiple Outcomes.}
The protocol designates EWMScore-P as the primary endpoint, Motion Quality and Trajectory Accuracy as mechanism-focused secondary endpoints, and Visual Quality and Interaction Quality as guardrails. It fixes component evaluators and aggregation before comparison and includes all specified outcomes regardless of significance. The paired intervals for secondary and guardrail outcomes are reported descriptively; we make no multiplicity-adjusted significance claims for those outcomes.

\paragraph{Matched-Run Requirements.}
The comparison is considered valid only when both variants share:
\begin{itemize}
    \item the exact parent-checkpoint hash and optimizer-reset policy;
    \item the same ordered training examples, augmentation decisions, update count, and random seed;
    \item identical sampler, initial noise, chunk boundaries, interpolation, and video encoding;
    \item the same evaluation manifest and evaluator version, with episode membership specified independently of the collection size; and
    \item retained per-episode metric files, logs, generated-video manifest, and checkpoint checksums.
\end{itemize}
A frozen parent checkpoint is a diagnostic rather than the primary baseline because it receives fewer updates. Likewise, the retained FlowWAM-FiveAges profile is external context, not a verified local matched control.

\paragraph{Artifact Provenance.}
A reproducible comparison links each training configuration and checkpoint to its evaluation inputs, generated outputs, and per-episode metrics. Data identifiers include source, task, and episode rather than numeric indices alone. External system profiles remain separate from locally reproduced comparisons; aggregate values alone do not yield episode-paired uncertainty estimates.

\subsection{Additional Diagnostics and Comparison Protocols}
\label{app:additional_ablations}

The following sections separate transport-input and temporal-reconstruction diagnostics from the matched objective comparison. Diagnostic datasets follow Appendix~\ref{app:dataset_protocol}; the matched TAWD protocol appears in Appendix~\ref{app:tawd_design_protocol}.

\subsubsection{KTL Transport Construction}
\label{app:ktl_transport_protocol}

Table~\appmainref{tab:ktl_design_ablation} reports a fixed-checkpoint diagnostic on 100 episodes from the separate three-view validation collection. All-view RAFT supplies guidance to the head, left-wrist, and right-wrist views; the control retains head-view guidance and replaces both wrist inputs with white placeholders. Each variant contains 300 frame-count-matched videos, with 4,800 sampled prediction--reference frame pairs in the reported evaluation. First-frame fidelity is reported separately because the initial RGB observation is a model condition.

$\operatorname{N\mbox{-}MAE}$ is normalized RGB mean absolute error, reported in $10^{-3}$ units. $\operatorname{Mot.\ Corr.}$ is the per-video Pearson correlation between sampled prediction and reference temporal-difference sequences. $\operatorname{Head}$, $\operatorname{Left}$, and $\operatorname{Right}$ report view-specific PSNR, while $\operatorname{First}$ and $\operatorname{Last}$ report boundary-frame PSNR. These are retained local video diagnostics, not official RoboTwin~2.0 scores.

\paragraph{Episode Coverage.}
Both generation inventories cover episodes 1--100 with one head-camera and two wrist-camera videos per episode. The three-view collection uses its own episode numbering rather than the Clean-50 split definition; identical numeric indices do not imply identical demonstrations.

\subsubsection{Matched TAWD Objective Comparison}
\label{app:tawd_design_protocol}

Table~\appmainref{tab:tawd_design_ablation} reports the executed matched comparison between $\beta=0$ and $\beta=1$.
The comparison fixes KTL and the aggregate future-RGB token weight, and changes only the spatial allocation of that weight.

\paragraph{Executed Weight Definitions.}
The uniform continuation uses $\beta=0$ and therefore $w_i=1$. The TAWD continuation uses $\beta=1$, giving $w_i=(1+N\rho_i^{\mathrm{tr}})/2$ and assigning half of the aggregate weight uniformly. Both variants have unit mean token weight and recover uniform weights under the zero-support fallback in Equation~\appmainref{eq:transport_distribution}.

\paragraph{Matched Comparison and Interpretation.}
TAWD improves EWMScore-P by 1.75 points (95\% paired bootstrap CI $[+1.10,+2.37]$), Trajectory Accuracy by 5.60 points, Interaction Quality by 4.20 points, and Motion Quality by 0.85 points. The Visual Quality interval crosses zero ($\Delta=+0.60$, CI $[-0.10,+1.20]$). These results apply to the matched 50-update continuation and do not measure the separate effect of KTL or longer training.

\subsubsection{Frame-Level Diagnostic Definitions}
\label{app:frame_diagnostics}

The measurements in Table~\ref{tab:main_frame_diagnostics} are complementary local diagnostics rather than newly proposed or official benchmark metrics. Let $Y_t\in[0,1]^{H\times W}$ denote the luma channel of frame $t$. $\operatorname{A\mbox{-}MAD}$ averages the pixel-wise absolute difference between adjacent $Y_t$; $\operatorname{CPF}$ reports the fraction of pixels whose adjacent-frame change exceeds the evaluator's fixed threshold; and $\operatorname{FL\mbox{-}MAD}$ applies the same absolute-difference statistic to the first and last frames. $\operatorname{NDP}$ is the fraction of adjacent pairs classified as near duplicates by a fixed frame-difference test. Frame differencing and cross-frame mean absolute differences have long been used in video analysis, including temporal-flicker evaluation \citep{huang2024vbench}.

For motion diagnostics, we estimate dense flow between adjacent frames with Farneb\"ack's two-frame method \citep{farneback2003}. $\operatorname{FB\ Mean}$ and $\operatorname{FB\ P95}$ are the mean and 95th percentile of the resulting flow magnitudes, while $\operatorname{Jerk}$ summarizes temporal variation across successive flow fields as a heuristic smoothness measure. Related differential-flow and jerk-based analyses precede this work \citep{shen2020blurry,takeda2018jerk}. These statistics characterize different failure modes: adjacent differences are sensitive to both motion and flicker, first-to-last difference measures endpoint change but not path quality, near-duplicate rate exposes repetition, and flow magnitude or jerk does not by itself establish correct action-conditioned motion. We therefore assign no universal optimization direction and use the diagnostics only to interpret the interpolation ablation.

\subsubsection{Temporal Reconstruction and Interpolation}

\begin{table*}[t]
\appendixtablestyle
\caption{\textbf{Interpolation Strategy.} Diagnostic on the same 16 episodes with fixed generated keyframes. Linear expansion uses linear frame interpolation; VFIMamba and RIFE HDv3 use learned interpolation.}
\begin{appendixtabular}{l YY YYY YYY}
\toprule
\multirow{2}{*}{$\operatorname{Method}$} & \multicolumn{2}{c}{$\operatorname{Appearance}$} & \multicolumn{3}{c}{$\operatorname{Motion}$} & \multicolumn{3}{c}{$\operatorname{Consistency}$} \\
\cmidrule(lr){2-3}\cmidrule(lr){4-6}\cmidrule(lr){7-9}
& $\operatorname{AES}\uparrow$ & $\operatorname{IMG}\uparrow$ & $\operatorname{DYN}\uparrow$ & $\operatorname{FL}\uparrow$ & $\operatorname{MS}\uparrow$ & $\operatorname{BG}\uparrow$ & $\operatorname{PS}\uparrow$ & $\operatorname{SC}\uparrow$ \\
\midrule
Linear expansion & \textbf{41.86} & 45.43 & 22.47 & 0.37 & 58.75 & 85.00 & 1.02 & 81.30 \\
VFIMamba (no TTA) & 40.30 & 45.50 & 41.65 & 8.37 & 63.43 & \textbf{90.50} & 54.22 & 86.97 \\
\rowcolor{red!5} RIFE HDv3 & 39.53 & \textbf{45.95} & \textbf{43.23} & \textbf{10.85} & \textbf{63.47} & 90.42 & \textbf{65.62} & \textbf{87.13} \\
\bottomrule
\end{appendixtabular}
\label{tab:main_interpolation_ablation}
\end{table*}

\begin{figure}[!htbp]
    \centering
    \begin{minipage}[c]{0.55\linewidth}
        \centering
        \includegraphics[width=\linewidth]{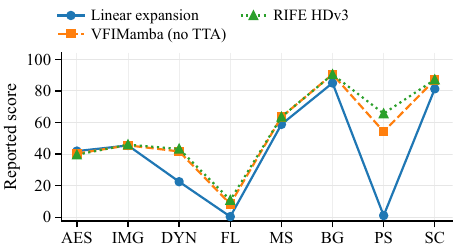}
    \end{minipage}\hfill
    \begin{minipage}[c]{0.42\linewidth}
        \small
        \caption{\small\raggedright\textbf{Interpolation diagnostics on 16 local episodes.} The methods expand the same generated keyframes; values match Table~\ref{tab:main_interpolation_ablation}. AES, IMG, DYN, FL, MS, BG, and SC follow Table~\ref{tab:full_metrics}; PS denotes Photometric Smoothness. This comparison evaluates temporal reconstruction with the generator fixed, rather than the isolated effect of TAWD.}
        \label{fig:interpolation_diagnostics}
    \end{minipage}
\end{figure}

\paragraph{Effects of Different Interpolators.}
This ablation tests whether learned interpolation improves motion and visual quality over linear expansion with the generated keyframes fixed (Table~\ref{tab:main_interpolation_ablation} and Figure~\ref{fig:interpolation_diagnostics}). RIFE improves seven of eight metrics over linear expansion and raises Dynamic Degree and Flow Score from 22.47 and 0.37 to 43.23 and 10.85, respectively, demonstrating stronger motion recovery despite a small aesthetic trade-off.

\paragraph{Effects of Temporal Reconstruction.}
Table~\ref{tab:main_temporal_reconstruction} compares temporal reconstruction on fixed generated outputs. Keyframe-6 retains every fourth frame of each 24-fps source and exports the sequence at 6 fps, providing a sparse-frame reference for the 24-fps reconstruction variants. It yields the highest Dynamic Degree and Flow Score under this coarser temporal sampling. Among the 24-fps variants, RIFE leads all three reported metrics, achieving Dynamic Degree of 43.23, Flow Score of 10.85, and Photometric Smoothness of 65.62. These results support RIFE for temporal reconstruction at the target frame rate.

\begin{table}[H]
    \centering
    \begin{minipage}[c]{0.55\linewidth}
        \appendixtablestyle
        \fontencoding{T1}\fontfamily{ptm}\fontsize{7.5}{9}\selectfont
        \renewcommand{\arraystretch}{1.05}
        \setlength{\tabcolsep}{2pt}
        \begin{appendixtabular}{l YYY}
        \toprule
        \multirow{2}{*}{$\operatorname{Temporal\ Variant}$} & \multicolumn{3}{c}{\shortstack{Local Motion\\Metrics (\%)}} \\
        \cmidrule(lr){2-4}
        & $\operatorname{DYN}\uparrow$ & $\operatorname{FL}\uparrow$ & $\operatorname{PS}\uparrow$ \\
        \midrule
        Keyframe-6 (6 fps) & 45.80 & 12.20 & 26.40 \\
\midrule
Linear expansion & 22.47 & 0.37 & 1.02 \\
Keyframe hold-4 & 33.09 & 9.69 & 13.78 \\
VFIMamba (no TTA) & 41.65 & 8.37 & 54.22 \\
\rowcolor{red!5} RIFE HDv3 & \textbf{43.23} & \textbf{10.85} & \textbf{65.62} \\
\bottomrule
\end{appendixtabular}
\end{minipage}\hfill
\begin{minipage}[c]{0.42\linewidth}
    \small
    \caption{\small\raggedright\textbf{Temporal Reconstruction.}
    Comparison on the same 16 local episodes with the generator fixed.
    Keyframe-6 reports the sparse generated keyframes directly at 6 fps,
    whereas the remaining variants produce 24-fps outputs.
    Bold marks the best results among the 24-fps reconstruction variants.
    $\operatorname{DYN}$, $\operatorname{FL}$, and $\operatorname{PS}$ denote
    Dynamic Degree, Flow Score, and Photometric Smoothness.}
    \label{tab:main_temporal_reconstruction}
\end{minipage}
\end{table}

\paragraph{Frame-Level Motion Analysis.}
This diagnostic tests whether interpolation changes intermediate motion and frame repetition without shifting the endpoints (Table~\ref{tab:main_frame_diagnostics}). RIFE increases mean optical-flow magnitude from 0.109 to 0.337 and reduces near-duplicate pairs from 0.588 to 0.563 while preserving the first-to-last difference, confirming that its gains arise from richer intermediate motion rather than endpoint drift.

\begin{table*}[t]
\appendixtablestyle
\caption{\textbf{Frame Diagnostics} for the 16-episode interpolation ablation. Definitions, interpretation, and metric provenance are provided in Appendix~\ref{app:frame_diagnostics}.}
\begin{appendixtabular}{l YYY c YYY}
\toprule
\multirow{2}{*}{$\operatorname{Method}$} & \multicolumn{3}{c}{$\operatorname{Pixel\ Difference}$} & \multicolumn{1}{c}{$\operatorname{Repetition}$} & \multicolumn{3}{c}{$\operatorname{Optical\ Flow}$} \\
\cmidrule(lr){2-4}\cmidrule(lr){5-5}\cmidrule(lr){6-8}
& $\operatorname{A\mbox{-}MAD}$ & $\operatorname{CPF}$ & $\operatorname{FL\mbox{-}MAD}$ & $\operatorname{NDP}$ & $\operatorname{FB\ Mean}$ & $\operatorname{FB\ P95}$ & $\operatorname{Jerk}$ \\
\midrule
Linear expansion & 0.006884 & 0.049167 & 0.096238 & 0.588374 & 0.109187 & 0.418630 & 0.026805 \\
VFIMamba (no TTA) & 0.008584 & 0.038254 & 0.096187 & 0.570205 & 0.324903 & 1.399471 & 0.078122 \\
\rowcolor{red!5} RIFE HDv3 & 0.008832 & 0.039097 & 0.096191 & 0.563069 & 0.337443 & 1.428315 & 0.075933 \\
\bottomrule
\end{appendixtabular}
\label{tab:main_frame_diagnostics}
\end{table*}

\paragraph{Scaling to Representative-100.}
This comparison tests whether RIFE's gains extend to interaction, perspective, and instruction following across 100 fixed episodes (Tables~\ref{tab:main_rep100_vlm} and~\ref{tab:main_rep100_outcomes}). RIFE improves Perspectivity by 0.40 points but reduces Interaction Quality and Instruction Following by 0.80 and 0.60 points, respectively, showing that interpolation gains do not transfer uniformly to semantic control.

\begin{table}[!htbp]
    \centering
    \begin{minipage}[c]{0.55\linewidth}
        \appendixtablestyle
        \fontencoding{T1}\fontfamily{ptm}\fontsize{7.5}{9}\selectfont
        \renewcommand{\arraystretch}{1.05}
        \setlength{\tabcolsep}{2pt}
        \begin{appendixtabular}{l YYY}
        \toprule
        \multirow{2}{*}{$\operatorname{Method}$} & \multicolumn{3}{c}{\shortstack{Paired Local VLM\\Metrics (\%)}} \\
        \cmidrule(lr){2-4}
        & $\operatorname{INT}\uparrow$ & $\operatorname{PER}\uparrow$ & $\operatorname{IF}\uparrow$ \\
        \midrule
        Linear expansion & \textbf{67.40} & 87.40 & \textbf{73.40} \\
        \rowcolor{red!5} RIFE HDv3 & 66.60 & \textbf{87.80} & 72.80 \\
        RIFE $-$ Linear (pp) & $-0.80$ & $+0.40$ & $-0.60$ \\
        \bottomrule
        \end{appendixtabular}
    \end{minipage}\hfill
    \begin{minipage}[c]{0.42\linewidth}
        \small
        \caption{\small\raggedright\textbf{Representative-100} paired comparison on identical episodes. $\operatorname{INT}$, $\operatorname{PER}$, and $\operatorname{IF}$ denote Interaction Quality, Perspectivity, and Instruction Following. Values are local VLM-judge percentages.}
        \label{tab:main_rep100_vlm}
    \end{minipage}
\end{table}

\begin{table}[!htbp]
    \centering
    \begin{minipage}[c]{0.55\linewidth}
        \appendixtablestyle
        \fontencoding{T1}\fontfamily{ptm}\fontsize{7.5}{9}\selectfont
        \renewcommand{\arraystretch}{1.05}
        \setlength{\tabcolsep}{2pt}
        \begin{appendixtabular}{l YYY}
        \toprule
        \multirow{2}{*}{$\operatorname{Metric}$} & \multicolumn{3}{c}{\shortstack{RIFE versus Linear\\(episodes)}} \\
        \cmidrule(lr){2-4}
        & $\operatorname{Improved}$ & $\operatorname{Tied}$ & $\operatorname{Regressed}$ \\
        \midrule
        Interaction Quality & 8 & 80 & 12 \\
        Perspectivity & 8 & 86 & 6 \\
        Instruction Following & 12 & 73 & 15 \\
        \bottomrule
        \end{appendixtabular}
    \end{minipage}\hfill
    \begin{minipage}[c]{0.42\linewidth}
        \small
        \caption{\small\raggedright\textbf{Paired Outcome} counts for RIFE versus linear expansion on Representative-100. Each row sums to 100; ties denote unchanged VLM-judge scores.}
        \label{tab:main_rep100_outcomes}
    \end{minipage}
\end{table}

\paragraph{Efficiency of Different Interpolators.}
This comparison tests whether RIFE reduces interpolation latency relative to VFIMamba under the same video-processing protocol (Table~\ref{tab:main_interpolation_efficiency}). RIFE reduces the mean per-episode time from 1168.24 to 6.02 seconds, yielding a $194.07\times$ speedup excluding the shared video-generation cost.

\begin{table}[!htbp]
    \centering
    \begin{minipage}[c]{0.55\linewidth}
        \appendixtablestyle
        \fontencoding{T1}\fontfamily{ptm}\fontsize{7.5}{9}\selectfont
        \renewcommand{\arraystretch}{1.05}
        \setlength{\tabcolsep}{2pt}
        \begin{appendixtabular}{l YYY}
        \toprule
        \multirow{2}{*}{$\operatorname{Method}$} & \multicolumn{2}{c}{$\operatorname{Measured\ Time\ (s)}$} & \multicolumn{1}{c}{\shortstack{Relative\\Efficiency}} \\
        \cmidrule(lr){2-3}\cmidrule(lr){4-4}
        & $\operatorname{Mean}\downarrow$ & $\operatorname{Total}\downarrow$ & $\operatorname{Speedup}\uparrow$ \\
        \midrule
        VFIMamba (no TTA) & 1168.24 & 18691.88 & 1.00$\times$ \\
        \rowcolor{red!5} RIFE HDv3 & \textbf{6.02} & \textbf{96.32} & \textbf{194.07}$\times$ \\
        \bottomrule
        \end{appendixtabular}
    \end{minipage}\hfill
    \begin{minipage}[c]{0.42\linewidth}
        \small
        \caption{\small\raggedright\textbf{Interpolation Efficiency} measured on 16 episodes. $\operatorname{Mean}$ is per-episode time, $\operatorname{Total}$ is summed accelerator time, and $\operatorname{Speedup}$ is relative to VFIMamba.}
        \label{tab:main_interpolation_efficiency}
    \end{minipage}
\end{table}

\paragraph{Inference-Time Flow-Condition Scale.}
Table~\ref{tab:condition_scale} compares inference-time flow-conditioning strengths on the same 16 episodes without changing the generator. Increasing the scale from 1.00 to 1.50 raises Dynamic Degree by 4.69\% and lowers Flow Score by 5.27\%, relative to scale 1.00. This diagnostic concerns inference-time conditioning strength, not the effect of training with TAWD.

\begin{table}[!htbp]
    \centering
    \begin{minipage}[c]{0.55\linewidth}
        \appendixtablestyle
        \fontencoding{T1}\fontfamily{ptm}\fontsize{7.5}{9}\selectfont
        \renewcommand{\arraystretch}{1.05}
        \setlength{\tabcolsep}{2pt}
        \begin{appendixtabular}{l YY}
        \toprule
        \multirow{2}{*}{$\operatorname{Flow\mbox{-}Condition\ Scale}$} & \multicolumn{2}{c}{\shortstack{Local Motion\\Diagnostics}} \\
        \cmidrule(lr){2-3}
        & $\operatorname{DYN}\uparrow$ & $\operatorname{FL}\uparrow$ \\
        \midrule
        \rowcolor{red!5} 1.00 & 0.224695 & \textbf{0.003716} \\
        1.25 & 0.227275 & 0.003652 \\
        1.50 & \textbf{0.235240} & 0.003520 \\
        \bottomrule
        \end{appendixtabular}
    \end{minipage}\hfill
    \begin{minipage}[c]{0.42\linewidth}
        \small
        \caption{\small\raggedright\textbf{Inference-Time Flow-Condition Scale} on 16 local episodes. $\operatorname{DYN}$ and $\operatorname{FL}$ denote Dynamic Degree and Flow Score. This is an auxiliary inference diagnostic rather than a KineWorld component ablation.}
        \label{tab:condition_scale}
    \end{minipage}
\end{table}

\paragraph{Transport-Construction Ablation.}
This ablation tests whether measured wrist-view guidance improves multi-view prediction over white wrist-flow placeholders (Table~\appmainref{tab:ktl_design_ablation}). All-view RAFT improves every reported reconstruction and temporal-agreement metric over the white-wrist-flow control, with the largest PSNR gains in the right-wrist and final-frame measurements. This supports the multi-view transport-construction choice within KTL but does not isolate the TAWD training objective.


\subsubsection{Retained Multi-View Component Profiles}
Table~\ref{tab:robotwin_multiview_components} reports all 19 component scores for the methods in Table~\appmainref{tab:robotwin_multiview_profiles}. NPSNR denotes normalized PSNR; VLM-I/II/III and VQA denote the four cross-view consistency assessments; IF, SA, and JEPA denote instruction following, semantic alignment, and JEPA similarity; INT and PER denote interaction quality and perspectivity; STATE denotes dynamic-state alignment; FL and TA denote flow score and trajectory accuracy; SC, BG, and PS denote subject consistency, background consistency, and photometric smoothness; IMG and AES denote image and aesthetic quality.

\begin{table}[H]
\appendixtablestyle
\fontencoding{T1}\fontfamily{ptm}\fontsize{8}{9.2}\selectfont
\renewcommand{\arraystretch}{0.95}
\caption{\textbf{Complete retained multi-view component profiles.} All seven methods and numerical entries are preserved from the supplied comparison; these are external aggregates, not verified RoboTwin~2.0 local re-evaluations. The normalized score scale and preference direction are unchanged; NPSNR is not PSNR in dB. Bold denotes the largest retained values at the reported precision, not a verified local ranking.}
\label{tab:robotwin_multiview_components}
\setlength{\tabcolsep}{0.7mm}
\begin{appendixtabular}{l YYYYYY}
\toprule
\multicolumn{7}{c}{A. Tri-view consistency components} \\
\midrule
$\operatorname{Method}$ & $\operatorname{NPSNR}\uparrow$ & $\operatorname{SSIM}\uparrow$ & $\operatorname{VLM-I}\uparrow$ & $\operatorname{VLM-II}\uparrow$ & $\operatorname{VLM-III}\uparrow$ & $\operatorname{VQA}\uparrow$ \\
\midrule
BWM & 73.43 & 87.47 & 85.11 & \textbf{83.94} & \textbf{95.31} & 65.94 \\
WoVR\_Plus & \textbf{74.62} & \textbf{87.91} & 83.79 & \textbf{83.94} & 94.84 & \textbf{69.11} \\
DreamDojo & 59.78 & 71.58 & 77.14 & 73.26 & 88.19 & 47.82 \\
Motus & 60.13 & 76.31 & 63.68 & 67.56 & 87.22 & 45.33 \\
Genie Envisioner & 58.74 & 74.44 & 69.40 & 57.13 & 77.93 & 36.72 \\
Ctrl-World & 46.39 & 64.62 & 46.44 & 63.69 & 82.46 & 40.90 \\
\rowcolor{red!5} KineWorld (Ours) & 43.15 & 67.23 & \textbf{91.70} & 73.72 & 82.93 & 39.41 \\
\bottomrule
\end{appendixtabular}

\par\vspace{6pt}
\begin{appendixtabular}{l YYYYY}
\toprule
\multicolumn{6}{c}{B. Task and physical components} \\
\midrule
$\operatorname{Method}$ & $\operatorname{IF}\uparrow$ & $\operatorname{SA}\uparrow$ & $\operatorname{JEPA}\uparrow$ & $\operatorname{INT}\uparrow$ & $\operatorname{PER}\uparrow$ \\
\midrule
BWM & \textbf{73.84} & 89.93 & \textbf{94.37} & 33.59 & 87.20 \\
WoVR\_Plus & 70.67 & \textbf{90.61} & 87.12 & 28.76 & 85.49 \\
DreamDojo & 49.72 & 88.28 & 30.73 & 28.28 & 59.39 \\
Motus & 52.28 & 86.54 & 10.24 & 25.00 & 44.20 \\
Genie Envisioner & 20.20 & 79.30 & 0.02 & 20.10 & 87.91 \\
Ctrl-World & 38.90 & 81.94 & 10.32 & 24.19 & 45.15 \\
\rowcolor{red!5} KineWorld (Ours) & 62.78 & 89.02 & 0.37 & \textbf{37.55} & \textbf{88.92} \\
\bottomrule
\end{appendixtabular}

\par\vspace{6pt}
\begin{appendixtabular}{l YYYYYYYY}
\toprule
\multicolumn{9}{c}{C. Motion, temporal, and visual components} \\
\midrule
$\operatorname{Method}$ & $\operatorname{FL}\uparrow$ & $\operatorname{TA}\uparrow$ & $\operatorname{STATE}\uparrow$ & $\operatorname{SC}\uparrow$ & $\operatorname{BG}\uparrow$ & $\operatorname{PS}\uparrow$ & $\operatorname{IMG}\uparrow$ & $\operatorname{AES}\uparrow$ \\
\midrule
BWM & 20.48 & 41.97 & \textbf{61.42} & 84.20 & 70.29 & 33.95 & \textbf{38.71} & 24.13 \\
WoVR\_Plus & \textbf{24.85} & \textbf{45.27} & 60.91 & 84.22 & 75.19 & 32.90 & 36.22 & \textbf{26.06} \\
DreamDojo & 6.52 & 16.61 & 60.76 & 77.37 & \textbf{89.29} & 15.86 & 24.45 & 17.58 \\
Motus & 0.80 & 13.77 & 59.31 & 77.37 & 1.00 & 1.31 & 16.08 & 16.44 \\
Genie Envisioner & 0.40 & 0.84 & 59.28 & \textbf{91.40} & 1.07 & 4.08 & 19.36 & 15.57 \\
Ctrl-World & 21.29 & 6.09 & 60.47 & 77.05 & 49.02 & 12.44 & 18.37 & 14.92 \\
\rowcolor{red!5} KineWorld (Ours) & 4.69 & 32.93 & 59.21 & 86.02 & 66.71 & \textbf{56.97} & 33.88 & 24.42 \\
\bottomrule
\end{appendixtabular}
\end{table}

\raggedbottom
\subsection{Additional Qualitative Analysis}
\label{app:qualitative_protocol}

\begin{figure*}[p]
    \centering
    \includegraphics[page=1,width=\textwidth,height=0.9\textheight]{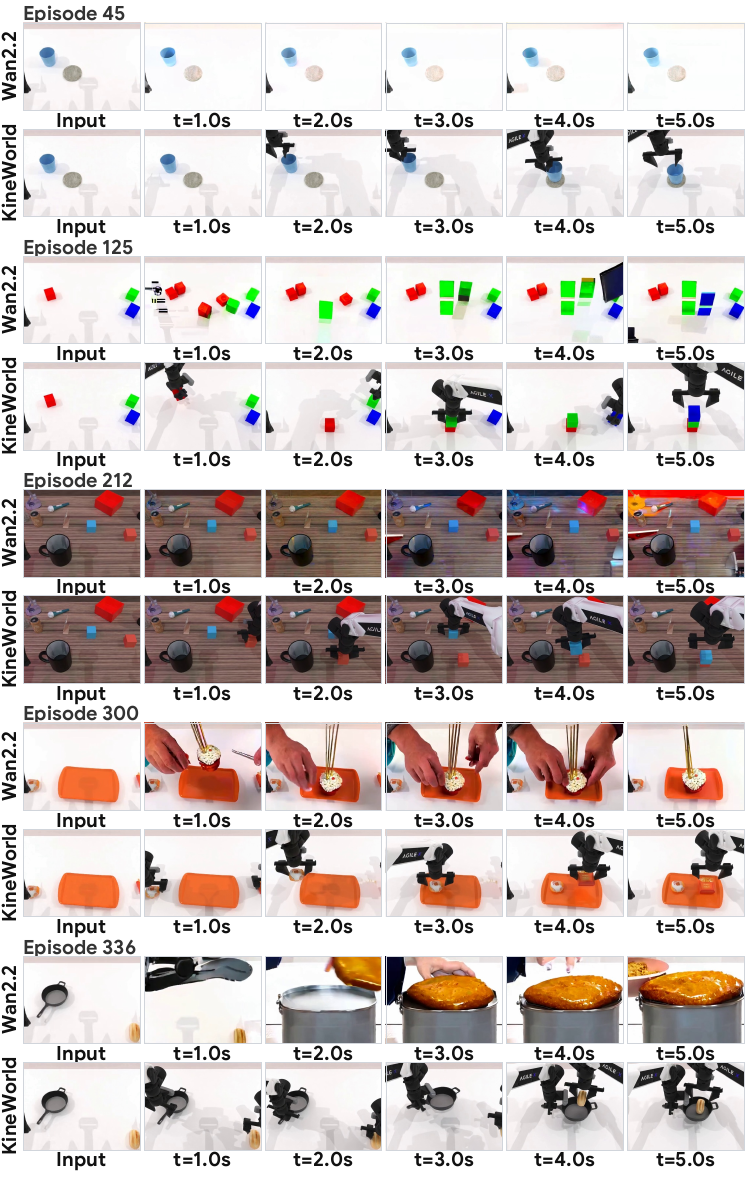}
    \caption{\textbf{Additional qualitative rollout comparisons (1/3).} Wan2.2 and KineWorld outputs for retained episode identifiers 45, 125, 212, 300, and 336. Each row shows the decoded input followed by five uniformly spaced rollout frames.}
    \label{fig:appendix_gallery_1}
\end{figure*}

\begin{figure*}[p]
    \centering
    \includegraphics[page=2,width=\textwidth,height=0.9\textheight]{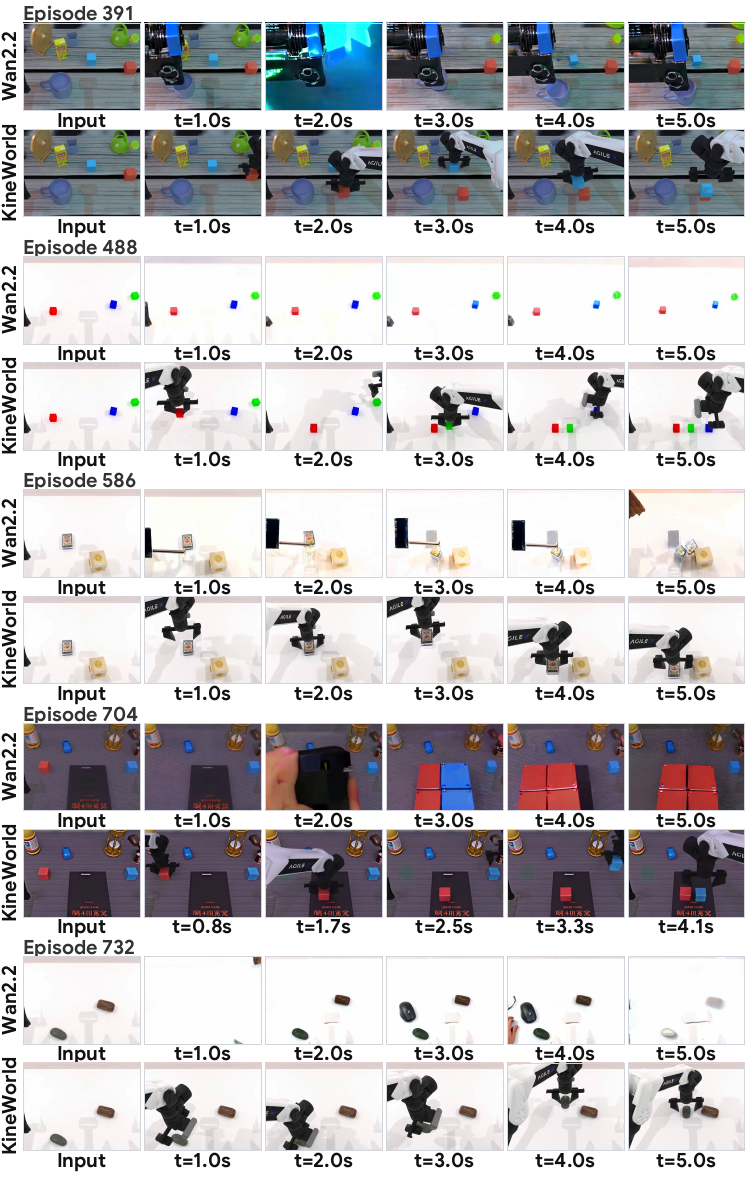}
    \caption{\textbf{Additional qualitative rollout comparisons (2/3).} Wan2.2 and KineWorld outputs for retained episode identifiers 391, 488, 586, 704, and 732, using the same deterministic frame-sampling rule as Figure~\ref{fig:appendix_gallery_1}.}
    \label{fig:appendix_gallery_2}
\end{figure*}

\begin{figure*}[p]
    \centering
    \includegraphics[page=3,width=\textwidth,height=0.9\textheight]{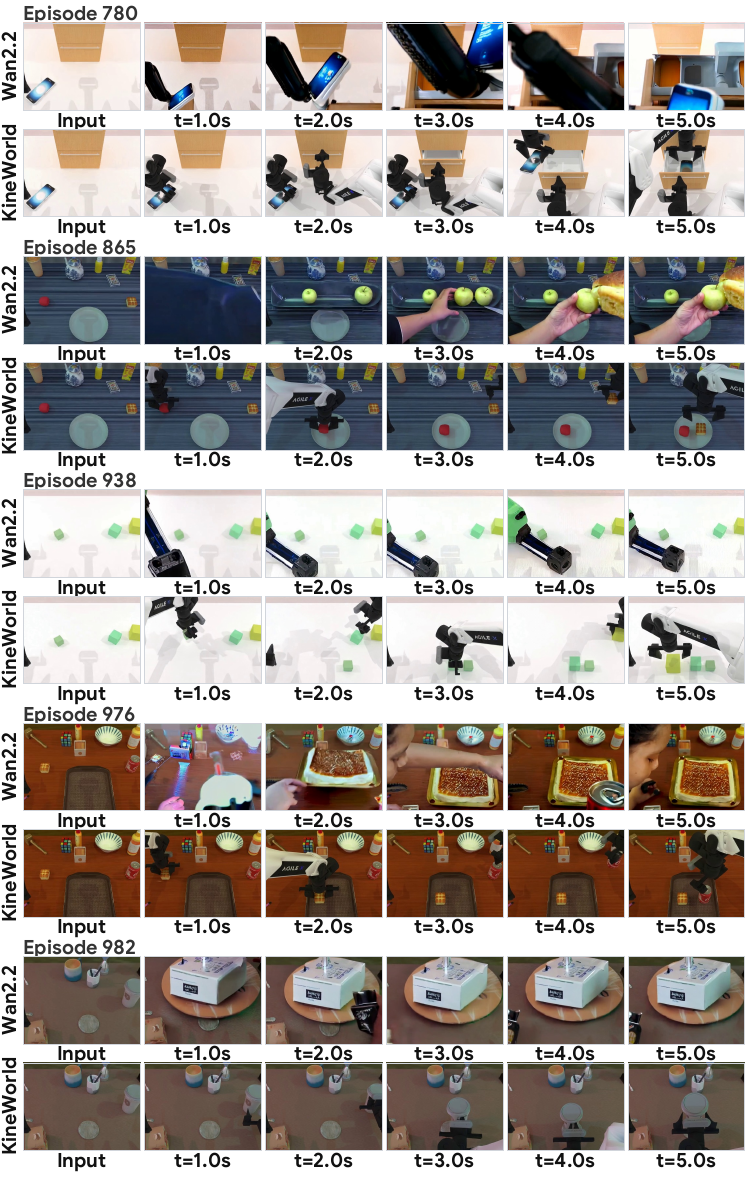}
    \caption{\textbf{Additional qualitative rollout comparisons (3/3).} Wan2.2 and KineWorld outputs for retained episode identifiers 780, 865, 938, 976, and 982, using the same deterministic frame-sampling rule as Figure~\ref{fig:appendix_gallery_1}.}
    \label{fig:appendix_gallery_3}
\end{figure*}

\begin{figure*}[p]
    \centering
    \includegraphics[page=1,width=\textwidth]{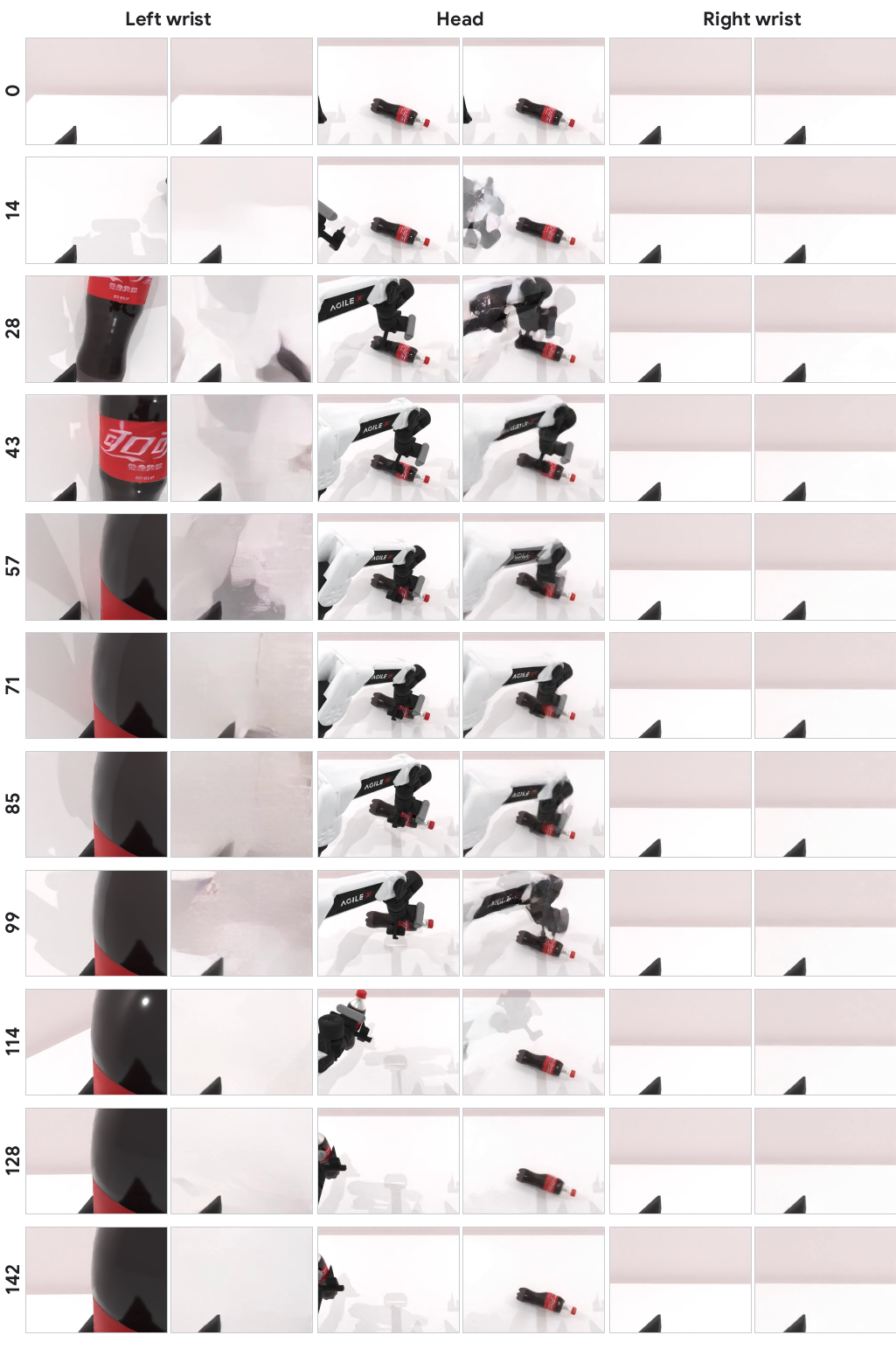}
    \caption{\textbf{Episode 1, bottle manipulation.} Columns are Left wrist, Head, and Right wrist; within each cell, the left image is GT and the right image is KineWorld. Eleven uniformly sampled frame indices are shown.}
    \label{fig:multiview_bottle}
\end{figure*}

\begin{figure*}[p]
    \centering
    \includegraphics[page=2,width=\textwidth]{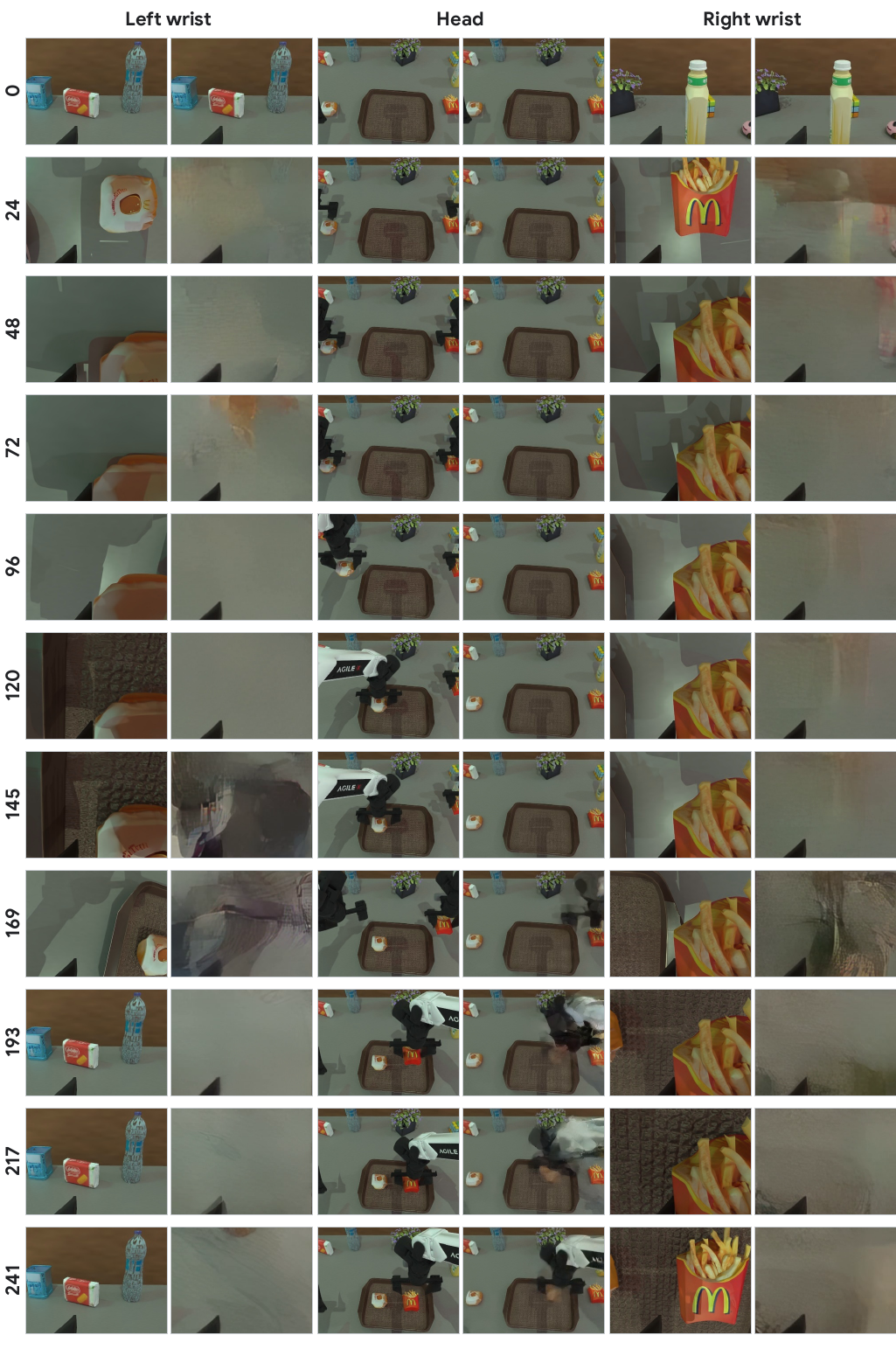}
    \caption{\textbf{Episode 50, bimanual food placement.} Columns are Left wrist, Head, and Right wrist; within each cell, the left image is GT and the right image is KineWorld. Eleven uniformly sampled frame indices are shown.}
    \label{fig:multiview_placement}
\end{figure*}

\begin{figure*}[p]
    \centering
    \includegraphics[page=3,width=\textwidth]{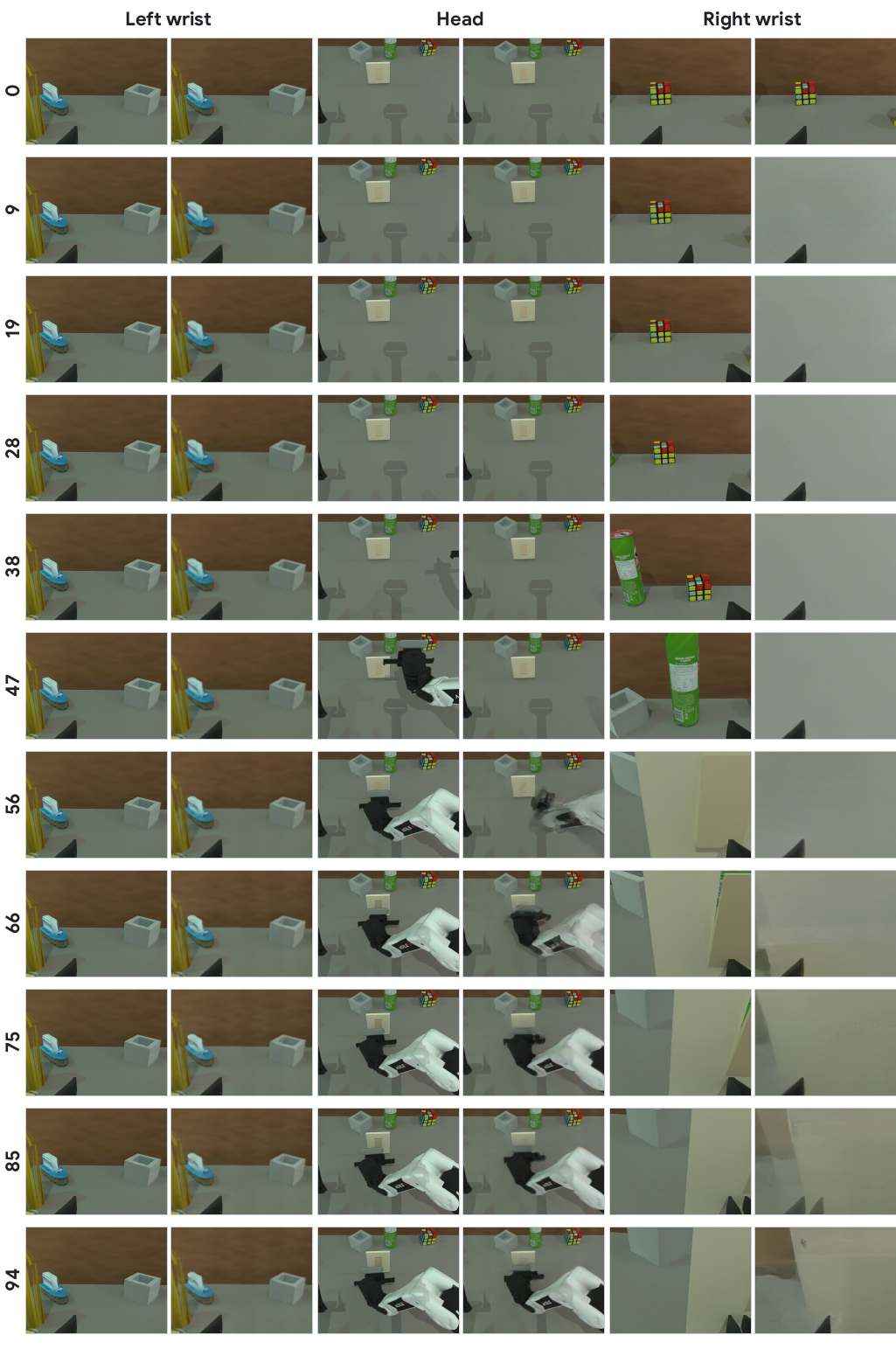}
    \caption{\textbf{Episode 100, switch interaction.} Columns are Left wrist, Head, and Right wrist; within each cell, the left image is GT and the right image is KineWorld. Eleven uniformly sampled frame indices are shown.}
    \label{fig:multiview_switch}
\end{figure*}

\paragraph{Visualization Protocol.}
Figures~\ref{fig:appendix_gallery_1}--\ref{fig:appendix_gallery_3} show 15 single-view episode pairs, each with its decoded input and five uniformly sampled future frames. Timestamps follow each video's encoded frame rate, without retiming. Figures~\ref{fig:multiview_bottle}--\ref{fig:multiview_switch} pair all-view-RAFT validation rollouts with ground truth at eleven shared frame indices across three cameras. These visualization collections are distinct from the Clean-50 held-out partition and do not constitute matched objective comparisons.

\paragraph{Single-View Comparisons.}
In Figures~\ref{fig:appendix_gallery_1}--\ref{fig:appendix_gallery_3}, Wan2.2 remains nearly static in episodes 45 and 488, whereas KineWorld depicts robot motion and object displacement. In episodes 300, 336, and 865, Wan2.2 introduces human hands or changes the objects and scene composition; KineWorld retains the robotic setting. These observations distinguish visible rollout behavior, but do not establish trajectory accuracy or task completion.

\paragraph{Multi-View Comparisons.}
Figures~\ref{fig:multiview_bottle}--\ref{fig:multiview_switch} expose view-dependent errors. Episode 1 retains the bottle on the table after the reference lift and misses its left-wrist close-up. Episode 50 shows divergent tray contents and reduced wrist-view object detail. Episode 100 preserves the static left-wrist scene but exhibits delayed head-view motion and missing right-wrist objects. Stable backgrounds therefore do not ensure accurate object transport across views. These validation diagnostics are separate from the scored multi-view test outputs.

\subsection{Limitations and Reproducibility}
\label{app:limitations}

\paragraph{Limitations.}
KineWorld relies on calibrated robot and camera models, and its transport support is sensitive to calibration errors, flow estimation errors, and occlusion. The support captures projected robot motion but does not explicitly model object motion or contact dynamics. Autoregressive rollout may accumulate appearance drift, while transport-focused supervision may trade visual fidelity for motion quality. The matched study evaluates TAWD over a 50-update continuation from a shared checkpoint; convergence and stability across training seeds remain to be assessed. Future work will investigate object-aware transport, calibration-robust conditioning, and cross-embodiment transfer.

\paragraph{Safety.}
Generated rollouts are predictive media rather than certified robot trajectories. They must not be used as evidence of collision-free or hardware-safe execution. Deployment requires independent state estimation, feasibility and collision checking, controller limits, emergency handling, and validation in the target simulator and robot system.

\paragraph{Reproducibility Specification.}
Reproducing the controlled comparison requires:
\begin{itemize}
    \item source code for rendering, support construction, training, sampling, and evaluation;
    \item pinned environment and dependency versions, including renderer, RAFT, and VAE revisions;
    \item data-split and episode manifests with checksums and license notes;
    \item parent and trained checkpoint hashes plus complete run configurations;
    \item external-source records and local or matched per-episode evaluations with bootstrap scripts;
    \item generated-video manifests linking each output to its input, seed, and checkpoint; and
    \item timing, memory, seed-variation, ablation, qualitative-selection, and failure-case reports.
\end{itemize}

\paragraph{Scope of Conclusions.}
The normalization and zero-motion properties follow from the objective definition. The reported system profiles and local diagnostics characterize their respective evaluation scopes; their interpretation is distinct from the causal effect of changing future-RGB supervision (Appendix~\ref{app:tawd_design_protocol}).

\ifVThreeAppendixStandalone
    \clearpage
    \bibliographystyle{iclr2027_conference}
    \bibliography{iclr2027_conference}
\fi
\end{document}

\end{document}